\documentclass{article}

\usepackage[preprint]{neurips_2026}

\usepackage[utf8]{inputenc}
\usepackage[T1]{fontenc}
\usepackage[colorlinks,linkcolor=black,citecolor=blue!70!black,urlcolor=blue!70!black]{hyperref}
\usepackage{url}
\usepackage{booktabs}
\usepackage{enumitem}
\usepackage{amsfonts}
\usepackage{amsmath}
\usepackage{amssymb}
\usepackage{nicefrac}
\usepackage{microtype}
\usepackage{xcolor}
\usepackage{graphicx}
\usepackage{multirow}
\usepackage{array}
\usepackage{bbm}
\usepackage{amsthm}
\usepackage[capitalize,noabbrev]{cleveref}
\usepackage{algorithm}
\usepackage{algpseudocode}
\usepackage{wrapfig}
\usepackage{tikz}
\usetikzlibrary{arrows.meta,calc,fit,positioning,shapes.geometric,backgrounds}
\usepackage[most]{tcolorbox}
\newtcolorbox{promptbox}{colback=blue!5!white,colframe=blue!75!black,fontupper=\footnotesize,fontlower=\footnotesize,boxrule=1pt,arc=2pt,width=\textwidth}

\theoremstyle{remark}

\usepackage{xspace}
\usepackage{makecell}
\usepackage{threeparttable}

\definecolor{myblue1}{HTML}{1F77B4}

\newif\ifrolecolor\rolecolortrue
\definecolor{challengercolor}{HTML}{E8873A}
\definecolor{solvercolor}{HTML}{2CA02C}
\definecolor{judgecolor}{HTML}{7B8FA1}
\definecolor{vaguecolor}{HTML}{B33F35}   
\definecolor{factcolor}{HTML}{0B536F}    
\definecolor{whycolor}{HTML}{7EC8A0}     
\newcommand{\vague}[1]{\protect\textcolor{vaguecolor}{\textbf{#1}}}
\newcommand{\fact}[1]{\protect\textcolor{factcolor}{\textbf{#1}}}
\newcommand{\why}[1]{\protect\textcolor{whycolor}{\textbf{#1}}}
\newcommand{\rub}[2]{\hangindent=1.2em\hangafter=1 #1.~#2}

\newcommand{\method}{SCOPE\xspace}
\newcommand{\polC}{\pi_C}
\newcommand{\polS}{\pi_S}
\newcommand{\corpus}{\mathcal{D}}

\newcommand{\CJ}{\mathcal{G}}
\newcommand{\bJ}{\mathbf{b}}

\title{\method: Self-Play via Co-Evolving Policies \\ for Open-Ended Tasks}

\author{%
  Wai-Chung Kwan\textsuperscript{1} \quad
  Aryo Pradipta Gema\textsuperscript{1} \quad
  Joshua Ong Jun Leang\textsuperscript{2} \quad
  Pasquale Minervini\textsuperscript{1,3} \\[0.6em]
  \textsuperscript{1}University of Edinburgh \quad
  \textsuperscript{2}Imperial College London \quad
  \textsuperscript{3}Miniml.AI \\[0.4em]
  \texttt{wkwan@ed.ac.uk}
}

\begin{document}

\maketitle

\vspace{-1em}
\makeatletter\if@preprint\makeatother
  \begin{center}
    {\fontsize{11pt}{13pt}\selectfont
      \href{https://github.com/EdinburghNLP/SCOPE}{\raisebox{-0.1em}{\includegraphics[height=1.2em]{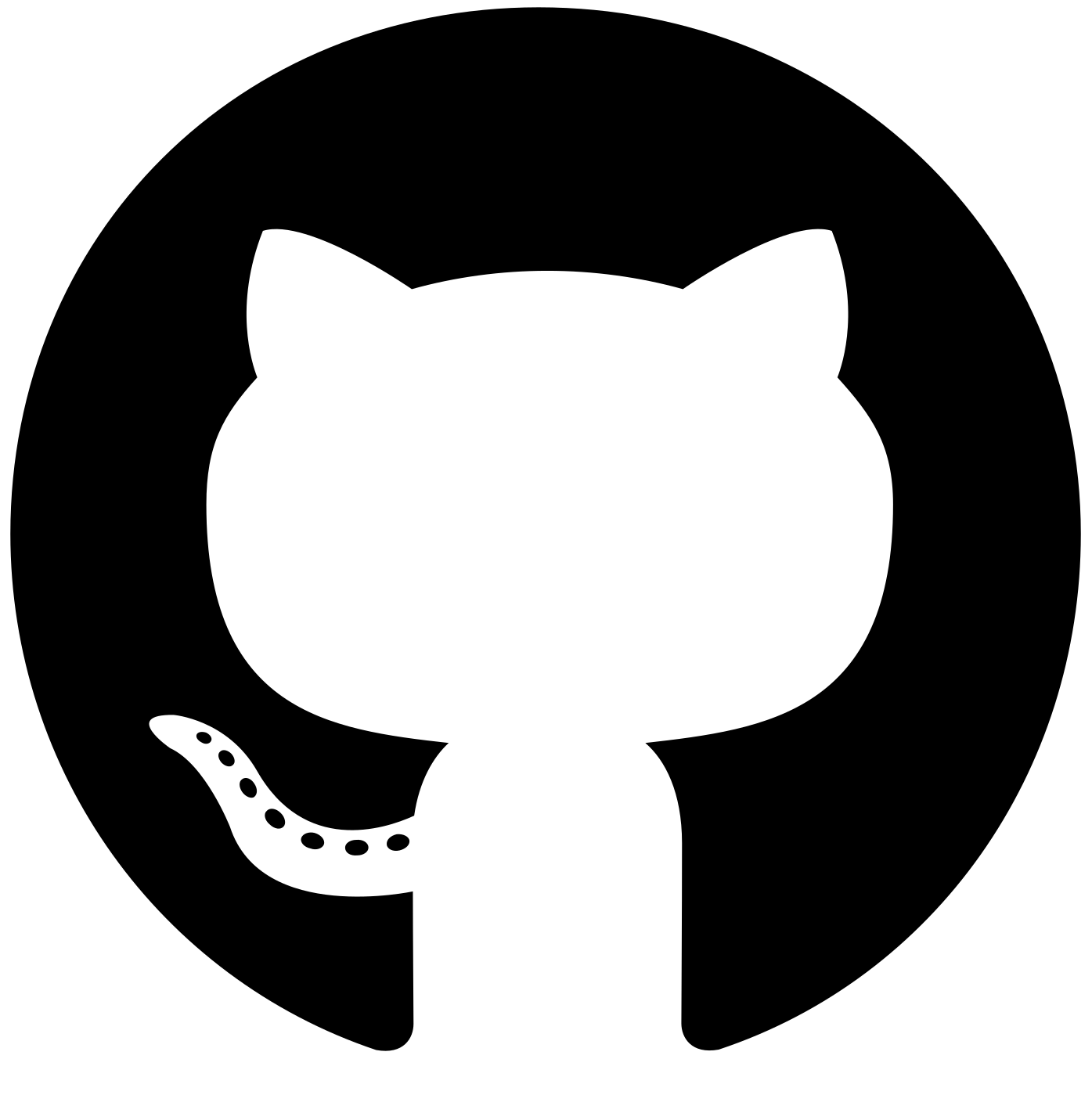}}~Code}
      \hspace{1.5em}
      \href{https://huggingface.co/collections/wckwan/scope}{\raisebox{-0.1em}{\includegraphics[height=1.2em]{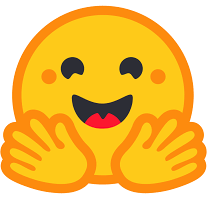}}~Models}
      \hspace{1.5em}
      \href{https://wandb.ai/cyruskwan/SCOPE/table}{\raisebox{-0.1em}{\includegraphics[height=1.2em]{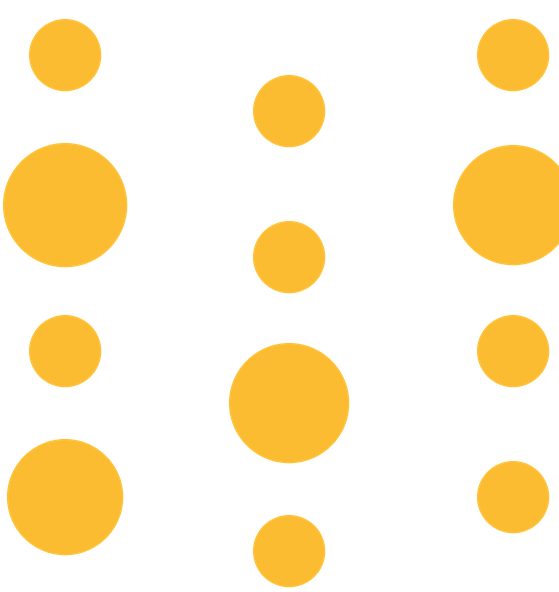}}\;\,Logs}}
  \end{center}
  \makeatletter\fi\makeatother
\vspace{1em}

\begin{abstract}
  Self-play can train language models without external supervision.
  However, existing methods require rule-checkable answers, leaving open-ended tasks dependent on curated prompts or frontier-model judges.
  We introduce \method, a data-free self-play framework for open-ended tasks that co-evolves two policies: \emph{a Challenger} that generates document-grounded tasks, and \emph{a Solver} that answers them through multi-turn retrieval.
  A frozen copy of the initial model serves as the \emph{self-judge}, which writes task-specific rubrics from the source document and grades Solver responses against them.
  Across three 7--8B instruction-tuned models (Qwen2.5, Qwen3, OLMo-3), \method improves open-ended performance by up to +10.4 points on eight benchmarks and matches or exceeds GRPO$_{\text{data}}$ trained on ${\sim}$9K curated prompts.
  Although trained only on open-ended tasks, \method also improves held-out short-form QA by up to +13.8 points on seven held-out benchmarks, surpassing GRPO$_{\text{data}}$ on all three models.
  Ablations show that co-evolving the Challenger is necessary to keep tasks near the Solver's frontier, that gains arise from improvements in both retrieval and synthesis with the relative contribution varying by task, and that rubric generation quality is the bottleneck for self-judging.
\end{abstract}

\section{Introduction} \label{sec:intro}

%
Reinforcement learning drives recent advances in language model capabilities~\citep{shao2024deepseekmath}, yet training for open-ended tasks such as deep research, complex QA, and writing still relies on human-curated prompts, reference answers, or frontier-model judges, tethering performance to human-level supervision~\citep{rar, drtulu, meta2025rpg}.
Existing approaches face two limits: human-curated supervision caps learning at human performance~\citep{silver2017mastering, hughes2024openendedness}, while stronger-model judges require access to a more capable model.
%

%
Self-play offers a path beyond both ceilings: models have achieved superhuman performance in games by learning entirely from self-generated experience~\citep{silver2016mastering, silver2017mastering,  brown2019superhuman, openai2019dota, vinyals2019grandmaster,baker2019emergent, meta2022diplomacy}.
Recent methods apply self-play to language models, co-evolving task-generating and task-solving policies without human-curated data~\citep{spice, azr, rzero, drzero,kwan2025opensir}.
Yet this line of work shares a fundamental dependency: tasks must have answers verifiable via string match \citep{drzero}, numerical equality \citep{rzero,spice,kwan2025opensir}, or code execution \citep{azr} to compute the reward signal that drives policy improvement.
%

Open-ended tasks admit many valid responses and lack a unique correct answer, placing them beyond the reach of rule-based verification.
Rubric-based evaluation offers an alternative, decomposing response quality into task-specific criteria that a language model can grade.
RaR~\citep{rar}, DR~Tulu~\citep{drtulu}, and RPG~\citep{meta2025rpg} show that rubric-based rewards can support open-ended RL; however, their use of curated prompts or frontier-model judges leaves unresolved the external-supervision bottleneck that motivates data-free self-play.

\begin{table}[t]
  \centering
  \footnotesize
  \begin{tabular}{l c c l}
    \toprule
    {\bf Method}                    & {\bf Open-Ended} & {\bf Data-Free} & {\bf Reward}          \\
    \midrule
    SPICE~\citep{spice}             & $\times$         & \checkmark      & Rule match            \\
    Dr.~Zero~\citep{drzero}         & $\times$         & \checkmark      & Rule match            \\
    R-Zero~\citep{rzero}            & $\times$         & \checkmark      & Rule match            \\
    Abs.~Zero~\citep{azr}           & $\times$         & \checkmark      & Code executor         \\
    OpenSIR~\citep{kwan2025opensir} & $\times$         & \checkmark      & Rule match            \\
    \midrule
    RaR~\citep{rar}                 & \checkmark       & $\times$        & Rubric (frontier LLM) \\
    DR~Tulu~\citep{drtulu}          & \checkmark       & $\times$        & Rubric (frontier LLM) \\
    RPG~\citep{meta2025rpg}         & \checkmark       & $\times$        & Rubric (self-judge)   \\
    \midrule
    \textbf{\method{}}              & \checkmark       & \checkmark      & Rubric (self-judge)   \\
    \bottomrule
  \end{tabular}
  \caption{\method{} is the first to extend data-free self-play to open-ended tasks. Prior self-play requires verifiable answers; rubric-based RL handles open-ended tasks but depends on curated prompts or frontier judges.}
  \label{tab:positioning}
\end{table}

We introduce \method (\textbf{S}elf-play via \textbf{Co}-evolving \textbf{P}olicies for \textbf{O}pen-\textbf{E}nded tasks), the first framework to extend data-free self-play to open-ended tasks (\cref{tab:positioning}).
\Cref{fig:method} provides an overview of our method.
\method co-evolves two policies from the same base model: a Challenger trained to generate open-ended, document-grounded tasks near the Solver's capability frontier, and a Solver trained to answer them through multi-turn retrieval.
A frozen copy of the base model serves as a fixed Judge, writing task-specific evaluation rubrics from the source document and scoring Solver responses against them.
At each iteration, the Challenger is rewarded for proposing moderately challenging document-grounded tasks for the current Solver.
The Solver is then rewarded for searching effectively and satisfying the Judge's rubrics on those tasks.
This creates the information asymmetry needed for sustained self-play~\citep{liu2026selfplay}: the Challenger and Judge condition on documents the Solver never sees, so it must recover what it needs through retrieval.

We evaluate \method{} on three 7--8B instruction-tuned models: Qwen2.5-7B~\citep{qwen2025qwen25}, Qwen3-8B~\citep{qwen2025qwen3}, and OLMo-3-7B~\citep{olmo2025olmo3}.
Across eight open-ended benchmarks, \method{} yields substantial improvements, with gains of up to +10.4 points on Qwen2.5-7B (24.4$\to$34.8; \cref{sec:exp:main}).
Despite using no curated prompts, it matches or exceeds GRPO$_{\text{data}}$ trained on ${\sim}$9K curated prompts.
This holds even on Qwen3-8B, the strongest base model tested and the setting with the least remaining headroom for improvement (\method{}/GRPO$_{\text{data}}$: 43.1 vs.\ 41.5).

\method{} also generalises more broadly than curated-prompt training.
Although trained exclusively on open-ended tasks, it surpasses GRPO$_{\text{data}}$ on short-form QA across all three model families, with gains of up to +13.8.
The contrast is largest on creative writing, a task without factual grounding, where \method{} surpasses GRPO$_{\text{data}}$ on all three backbones by up to +7.4, while GRPO$_{\text{data}}$ falls below the base model on two of them.
We find that co-evolution is essential: a frozen Challenger fails to sustain improvement beyond the first iteration.
The quality gates and cosine length penalty each prevent a distinct reward-hacking failure.
We further show that the quality of rubric generation, not grading capacity, is the bottleneck for self-judging (\cref{sec:analysis:judge}).
Performance gains arise from improvements in both retrieval and synthesis, and the relative importance of each depends on the task.
Together, these results show that data-free self-play produces transferable gains on open-ended tasks, extending self-improvement beyond the verifiable-answer domains to which it was previously confined.

\section{Background} \label{sec:background}
We assume access to an unlabelled document corpus $\corpus$ and a pretrained language model $\pi$.
Given a task $q$, $\pi_\theta$ generates $G$ rollouts $o_{1:G}$, each performing multi-turn retrieval over $\corpus$ to gather an evidence set $\mathbf{d}$ of supporting documents; an answer $a_i$ is parsed from each rollout $o_i$.
A judge $J$ evaluates each rollout against task-specific rubrics $\CJ = \{(c_k,\, w_k)\}_{k=1}^{K_q}$, where each $c_k$ is a natural-language criterion and $w_k \geq 0$ is its importance weight~\citep{rar, drtulu, meta2025rpg}.
For each criterion, the judge produces a binary verdict $b_k \in \{0,1\}$, collected into the grade vector $\bJ(o,\, q,\, \CJ) \in \{0,1\}^{K_q}$.
The rubric reward aggregates these verdicts as a weighted average:
\begin{equation} \label{eq:rubric_score}
  g(o,\, \CJ) = \frac{1}{\sum_{k=1}^{K_q} w_k} \sum_{k=1}^{K_q} w_k\, b_k.
\end{equation}
The overall reward combines multiple components $R(o) = \sum_j \lambda_j\, r_j(o)$, where $\lambda_j \geq 0$ are weights.
We optimise $\pi_\theta$ with group relative policy optimisation
(GRPO)~\citep{shao2024deepseekmath}, which estimates advantages from
within-group reward statistics without a learned value function.
For each prompt $x$, GRPO samples $G$ rollouts $o_{1:G}$ from $\pi_\theta$ and computes group-relative advantages:
\begin{equation} \label{eq:advantage}
  A_i = \frac{R(o_i) - \operatorname{mean}(R(o_{1:G}))} {\operatorname{std}(R(o_{1:G}))}.
\end{equation}
The policy is updated by maximising a clipped surrogate objective with KL
regularisation against a reference policy $\pi_{\mathrm{ref}}$:
\begin{equation} \label{eq:grpo}
  \mathcal{J}(\theta)
  = \mathbb{E}\!\left[\frac{1}{G}\sum_{i=1}^{G}
    \min\!\bigl(\rho_i\, A_i,\;
    \operatorname{clip}(\rho_i,\,1{-}\epsilon,\,1{+}\epsilon)\,A_i
    \bigr)\right]
  - \beta\,\mathbb{D}_{\mathrm{KL}}\!\bigl(\pi_\theta \,\|\,
  \pi_{\mathrm{ref}}\bigr),
\end{equation}
\noindent where $\rho_i = \pi_\theta(o_i \mid x) / \pi_{\theta_{\mathrm{old}}}(o_i \mid x)$ is the importance-sampling ratio, $\epsilon$ is the clipping range, and $\beta$ controls KL regularisation strength.

\section{\method} \label{sec:method}

\begin{figure*}[t]
  \centering
  \includegraphics[width=.95\textwidth]{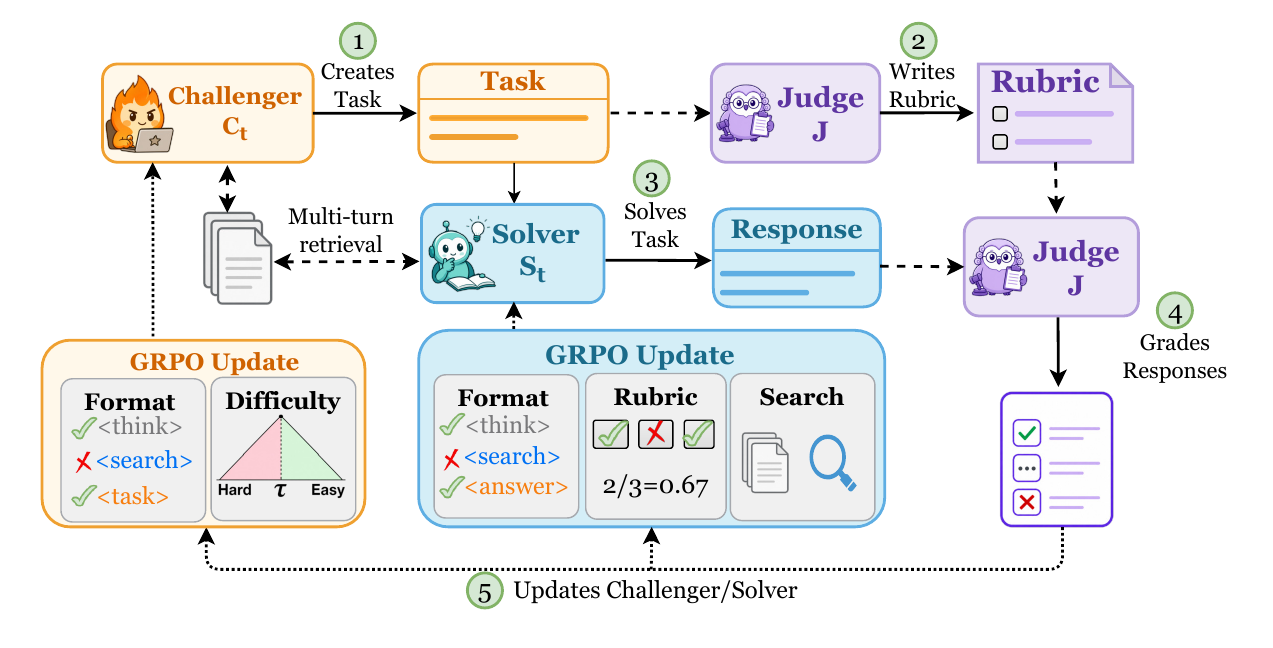}
  \caption{Overview of \method. From the same base model, \method trains a Challenger to generate document-grounded open-ended tasks from corpus documents $d$, and a Solver to answer them through multi-turn retrieval. A fixed Judge derives task-specific rubrics from $d$ and grades Solver responses. Each iteration alternates between \textbf{(1)}~Challenger training against the current Solver and \textbf{(2)}~Solver training on filtered Challenger-generated tasks with rubric-based rewards (details in \cref{sec:method:cycle}).}
  \label{fig:method}
\end{figure*}
\method organises self-play around three roles, all initialised from the same base model $M_0$ (\cref{fig:method}):
\begin{itemize}[leftmargin=*]
  \item \textbf{Challenger} $C_t$ (evolving): generates document-grounded tasks via multi-turn retrieval, trained to produce tasks at the Solver's capability frontier;
  \item \textbf{Solver} $S_t$ (evolving): answers tasks via multi-turn retrieval-augmented generation, trained to maximise rubric scores;
  \item \textbf{Judge} $J = M_0$ (fixed): generates task-specific evaluation rubrics from the source document, scores Solver responses against them, and applies quality gates to Challenger-generated tasks.
\end{itemize}
We denote the policies of $C_t$ and $S_t$ as $\polC$ and $\polS$, respectively; only these two are updated via GRPO~\citep{shao2024deepseekmath}, while $J$ remains fixed at $M_0$ throughout training.
We describe how training iterates between the two evolving roles and justify why this process sustains self-improvement in \cref{sec:method:sustainability}.

\subsection{Self-Play Training Loop}
\label{sec:method:cycle}

Each iteration of the self-play loop (\cref{alg:scope}) consists of two stages: Challenger training and Solver training.

\paragraph{Stage 1: Train Challenger.}
For a source document $d \in \corpus$ and a specified task type, the Challenger produces a multi-turn rollout $o^C$. During this rollout, it retrieves additional information (evidence set $\mathbf{d}$) and generates a candidate task $q$ (extracted from $o^C$) designed to require multi-turn retrieval to answer.
The Judge $J$ first applies a binary quality gate $\mathrm{QG}(q, \mathbf{d}) \in \{0,1\}$, which equals $1$ if all checks pass:
(1)~entity identifiability---the task references unambiguous entities that the Solver can locate via retrieval;
(2)~source relevance---the task is grounded in the provided document.
We find that without these gates, the Challenger degenerates to producing generic tasks unrelated to the source document to exploit the reward.
For each task that passes, $J$ generates rubrics $\CJ$. The previous Solver $S_{t-1}$ then generates $N$ multi-turn rollouts $o^S_{1:N}$, which $J$ grades against the rubrics.
We compute the rubric score $g(o^S, \CJ)$ as in \cref{eq:rubric_score}, then average the $N$ grades to obtain the group mean $\bar{g}$.
The Challenger reward combines format compliance with task difficulty:
\begin{equation}
  R_C(o^C, \mathbf{d}) = \lambda_{\mathrm{fmt}} \cdot r_{\mathrm{fmt}}(o^C) + \lambda_{\mathrm{diff}} \cdot \mathbbm{1}[\mathrm{QG}(q, \mathbf{d})] \cdot f_{\mathrm{diff}}(\bar{g};\,\tau),
  \label{eq:challenger_reward}
\end{equation}
where $r_{\mathrm{fmt}}(\cdot)$ is a role-specific format reward---$r_{\mathrm{fmt}}(o^C)$ checks the Challenger's think/search/task structure and $r_{\mathrm{fmt}}(o^S)$ checks the Solver's think/tool/answer structure.
The difficulty reward $f_{\mathrm{diff}}$ targets tasks at the Solver's capability frontier by comparing $\bar{g}$ to a target difficulty $\tau \in (0,1)$; it peaks where feedback variance $\bar{g}(1{-}\bar{g})$ is maximised:
\begin{equation}
  f_{\mathrm{diff}}(\bar{g};\,\tau) = \max\!\Bigl(0,\; 1 - \frac{|\bar{g} - \tau|}{\min(\tau,\, 1{-}\tau)}\Bigr),
  \label{eq:diff}
\end{equation}
with $\tau = \tfrac{1}{2}$, so the reward peaks when the Solver achieves $\bar{g} = \tfrac{1}{2}$ and vanishes at the extremes.
The Challenger is optimised with GRPO (\cref{eq:grpo}) to maximise $\mathbb{E}[R_C]$, using $\pi_{\mathrm{ref}} = M_0$ as the reference policy.

\paragraph{Stage 2: Train Solver.}
The trained Challenger $C_t$ first generates candidate tasks over $\corpus$.
Each candidate is evaluated: the Judge applies quality gates $\mathrm{QG}(q, \mathbf{d})$, generates rubrics $\CJ$, the previous Solver $S_{t-1}$ produces rollouts, and the Judge grades them.
We discard tasks that fail the quality gates or whose mean rubric score $\bar{g}$ falls outside a difficulty window $[\ell, u]$ ($\ell = 0.2$, $u = 0.8$), retaining only moderate-difficulty tasks.
We then optimise the Solver $S_t$ with GRPO (\cref{eq:grpo}) on these filtered tasks, again using $\pi_{\mathrm{ref}} = M_0$.
The Solver reward combines a length-controlled rubric score with format and search rewards:
\begin{equation}
  R_S(o^S, q, \mathbf{d}) = \lambda_{\mathrm{acc}} \cdot L(a) \cdot g(o^S, \CJ) + \lambda_{\mathrm{fmt}} \cdot r_{\mathrm{fmt}}(o^S) + \lambda_{\mathrm{search}} \cdot r_{\mathrm{search}}(o^S),
  \label{eq:solver_reward}
\end{equation}
where $a$ is the answer extracted from rollout $o^S$, $g(o^S, \CJ) \in [0, 1]$ is the rubric reward, $r_{\mathrm{fmt}}(o^S)$ is the format reward, and $r_{\mathrm{search}}(o^S)$ rewards search tool usage.
Without length control, the Solver produces increasingly long responses because the rubric judge tends to score longer outputs more favourably~\citep{rar, rubricanchors, scaleai2025researchrubrics}.
$L(a)$ counteracts length inflation with a cosine penalty:
\begin{equation}
  L(a) = \begin{cases}
    1                                                                                                                                                                   & \text{if } |a| \leq \ell_{\mathrm{soft}}                     \\[4pt]
    \ell_{\min} + \frac{1 - \ell_{\min}}{2}\left(1 + \cos\!\left(\pi \cdot \frac{|a| - \ell_{\mathrm{soft}}}{\ell_{\mathrm{hard}} - \ell_{\mathrm{soft}}}\right)\right) & \text{if } \ell_{\mathrm{soft}} < |a| < \ell_{\mathrm{hard}} \\[4pt]
    \ell_{\min}                                                                                                                                                         & \text{if } |a| \geq \ell_{\mathrm{hard}}
  \end{cases}
  \label{eq:length_penalty}
\end{equation}
where $|a|$ is the answer length in tokens, $\ell_{\mathrm{soft}}$ and $\ell_{\mathrm{hard}}$ are soft and hard token limits, and $\ell_{\min}$ is a floor value.
Intuitively, $L$ applies no penalty below $\ell_{\mathrm{soft}}$, then smoothly decays the score multiplier to a near-zero floor at $\ell_{\mathrm{hard}}$, discouraging the model from inflating answers beyond the soft limit without introducing a sharp reward discontinuity.

\subsection{Sustainable Self-Improvement}
\label{sec:method:sustainability}

We justify \method by showing it sustains self-improvement across iterations.
Following \citet{liu2026selfplay}, sustained self-play requires exposing \emph{learnable} information to the Solver---not merely generating more data.
Let $d \sim p_{\corpus}$ denote a source document, $q \sim C_t(\cdot \mid d)$ the task generated by the Challenger, and $\CJ \sim J(\cdot \mid q, d)$ the evaluation rubric produced by the Judge.
\method creates learnable information through document grounding: the Challenger and Judge observe $d$ when constructing tasks and rubrics, while the Solver sees only $q$ and must close the resulting gap through retrieval.
Two conditions determine whether the pipeline exposes learnable information.
First, \emph{task grounding}: the task must carry information about the source document that the Solver cannot answer from parametric knowledge alone.
We quantify this as the task--document mutual information:
\begin{equation} \label{eq:task_document_mi}
  I(q;\,d)
  =
  \mathbb{E}_{\begin{subarray}{l}d \sim p_{\corpus},\\ q \sim C_t(\cdot \mid d)\end{subarray}}
  \left[
    \log \frac{C_t(q \mid d)}{p_t(q)}
    \right] > 0,
  \quad \text{with} \ \ \
  p_t(q)=\mathbb{E}_{d \sim p_{\corpus}}\!\left[C_t(q \mid d)\right].
\end{equation}
Second, a \emph{hidden-rubric gap}: rubrics must encode document-specific criteria beyond what the task reveals, so that even after observing the task, the Solver's evaluation depends on information it does not yet possess.
We quantify this as the conditional mutual information between the rubric and the source document given the task:
\begin{equation} \label{eq:hidden_rubric_gap}
  I(\CJ;\,d \mid q)
  = H(\CJ \mid q) - H(\CJ \mid q, d) > 0.
\end{equation}
This quantity directly measures the learnable information in the pipeline: the rubric content that depends on the document beyond what the task alone reveals.
Task grounding ensures there is information the Solver lacks; the hidden-rubric gap ensures that precisely this information determines the learning signal.
As the Solver improves, it closes the hidden-rubric gap on current tasks and the learning signal saturates. The Challenger must therefore co-evolve to propose tasks where the gap remains open, sustaining the self-improvement loop.

\section{Experiments}
\label{sec:experiments}

\subsection{Setup}
\label{sec:exp:setup}

\paragraph{Training Details.}
We train and evaluate three instruction-tuned models (Qwen2.5-7B-Instruct~\citep{qwen2025qwen25}, Qwen3-8B~\citep{qwen2025qwen3}, and OLMo-3-7B-Instruct~\citep{olmo2025olmo3}). Self-play is most relevant for post-trained models because they have already absorbed human-curated data and require self-generated signal to improve further~\citep{kwan2025opensir}.
We train each model with GRPO~\citep{shao2024deepseekmath} for 3 iterations of 20 steps per role per iteration.
We sample training documents from English Wikipedia.
The Challenger generates four task types: long-form QA, summarisation, planning, and writing.
We use uniform rubric weights ($w_k = 1$ in \cref{eq:rubric_score}); \cref{tab:app_hyperparams} lists reward weights and length-penalty thresholds.
We report the last checkpoint of each iteration.

\paragraph{Baseline.}
We compare against GRPO$_{\text{data}}$, which trains each base model with GRPO on DR~Tulu's ${\sim}$9K curated prompts for open-ended tasks~\citep{drtulu} (5K from SearchArena/OpenScholar, 4K rubric-annotated prompts from RaR~\citep{rar}).
GRPO$_{\text{data}}$ uses the same base-model Judge as \method{}; the only difference is its reliance on externally curated prompts with frontier-model rubrics, versus \method{}'s fully self-generated tasks and rubrics.
This comparison tests whether data-free self-play can match training with curated data.

\paragraph{Benchmarks.}
We evaluate on eight open-ended benchmarks: deep research (DRB-RACE~\citep{du2025deepresearch}, ResearchRubrics~\citep{scaleai2025researchrubrics}), scholarly QA (ResearchQA~\citep{li2025researchqa}, SQA-CS-V2~\citep{allenai2025astabench}), planning (ResearchPlanGen~\citep{meta2025rpg}), user assistance (HealthBench~\citep{openai2025healthbench}, WildBench~\citep{lin2024wildbench}), and creative writing (Arena-Hard-CW~\citep{li2024arenahard}).

\looseness-1
To assess whether these gains generalise beyond open-ended tasks, we additionally evaluate on seven \emph{short-form QA} benchmarks: general QA (NQ~\citep{kwiatkowski2019natural}, TriviaQA~\citep{joshi2017triviaqa}, PopQA~\citep{mallen2023popqa}) and multi-hop QA (HotpotQA~\citep{yang2018hotpotqa}, 2WikiMultiHopQA~\citep{ho2020constructing}, MuSiQue~\citep{trivedi2022musique}, Bamboogle~\citep{press2023measuring}).
Following~\citep{drtulu}, we subsample benchmarks exceeding 1{,}000 examples to 1{,}000.

\paragraph{Evaluation Details.}
We score open-ended benchmarks with \texttt{gpt-5.4-mini} at $T{=}1.0$ and \texttt{reasoning\_effort=high}; for short-form QA we use \texttt{gpt-5-mini} at $T{=}1.0$.
Following prior work~\citep{drtulu, li2025webthinker, wei2024measuring}, we use LLM-as-judge rather than exact match for short-form QA\@.
Each task allows up to 5 search turns within a 24{,}576-token context window and 16{,}384-token answer budget; each search call retrieves up to 3 documents truncated to 500 tokens combined.
Qwen3-8B decodes at $T{=}1.0$ to avoid repetition loops; Qwen2.5-7B and OLMo-3-7B decode greedily.

\subsection{Main Results}
\label{sec:exp:main}

\begin{table}[t]
  \centering
  \footnotesize
  \setlength{\tabcolsep}{3.2pt}
  \begin{tabular}{l cc cc c cc c c}
    \toprule
                               & \multicolumn{2}{c}{Deep Research} & \multicolumn{2}{c}{Scholarly QA} & Planning      & \multicolumn{2}{c}{User Assist.} & Creative      &                                                               \\
    \cmidrule(lr){2-3} \cmidrule(lr){4-5} \cmidrule(lr){6-6} \cmidrule(lr){7-8} \cmidrule(lr){9-9}
    Model                      & DRB                               & Rubrics                          & ResQA         & SQAv2                            & ResPlan       & HealthB.      & WildB.        & AH-CW         & Avg.          \\
    \midrule
    Qwen2.5-7B                 & 33.2                              & 13.1                             & 32.0          & 25.8                             & 32.4          & 16.1          & 33.8          & 8.5           & 24.4          \\
    \quad GRPO$_{\text{data}}$ & 52.1                              & 24.4                             & 46.4          & \textbf{33.3}                    & 43.1          & \textbf{20.7} & 35.4          & 11.9          & 33.4          \\
    \quad \method\ iter1       & 47.5                              & 19.8                             & 46.0          & 31.0                             & 40.5          & 17.2          & 34.6          & 10.6          & 30.9          \\
    \quad \method\ iter2       & 50.0                              & 24.3                             & 48.5          & 31.0                             & 43.6          & 19.4          & \textbf{35.7} & 11.5          & 33.0          \\
    \quad \method\ iter3       & \textbf{53.2}                     & \textbf{27.3}                    & \textbf{50.5} & 32.8                             & \textbf{45.4} & 20.1          & 35.0          & \textbf{13.9} & \textbf{34.8} \\
    \midrule
    Qwen3-8B                   & 49.2                              & 24.2                             & 40.0          & 37.5                             & 53.9          & 25.9          & 45.7          & 24.9          & 37.7          \\
    \quad GRPO$_{\text{data}}$ & 56.8                              & \textbf{33.1}                    & 51.2          & 40.4                             & 57.0          & 25.1          & \textbf{47.0} & 21.0          & 41.5          \\
    \quad \method\ iter1       & 51.3                              & 25.8                             & 47.8          & 38.5                             & 55.3          & 26.2          & 46.2          & 26.5          & 39.7          \\
    \quad \method\ iter2       & 55.5                              & 30.1                             & 51.6          & 39.8                             & 56.8          & 26.8          & 46.9          & 27.6          & 41.9          \\
    \quad \method\ iter3       & \textbf{57.6}                     & 31.3                             & \textbf{53.7} & \textbf{41.8}                    & \textbf{57.7} & \textbf{28.1} & 46.1          & \textbf{28.4} & \textbf{43.1} \\
    \midrule
    OLMo-3-7B                  & 40.3                              & 19.0                             & 35.9          & 33.0                             & 51.1          & 16.2          & 35.5          & 14.8          & 30.7          \\
    \quad GRPO$_{\text{data}}$ & \textbf{50.7}                     & \textbf{32.8}                    & \textbf{52.3} & 37.8                             & 58.8          & \textbf{23.9} & \textbf{42.7} & 12.8          & \textbf{39.0} \\
    \quad \method\ iter1       & 45.4                              & 27.2                             & 44.1          & 34.8                             & 57.7          & 19.5          & 41.4          & 16.8          & 35.8          \\
    \quad \method\ iter2       & 47.7                              & 31.3                             & 48.3          & \textbf{38.0}                    & 58.8          & 19.9          & 41.2          & 18.0          & 37.9          \\
    \quad \method\ iter3       & 47.9                              & 31.6                             & 51.1          & 37.3                             & \textbf{59.5} & 21.2          & 40.9          & \textbf{18.7} & 38.5          \\
    \bottomrule
  \end{tabular}
  \caption{\textbf{Open-ended benchmark results.} Average scores improve monotonically across iterations for all three models. Though fully data-free, \method{} is comparable to GRPO$_{\text{data}}$ overall while improving creative writing where GRPO$_{\text{data}}$ regresses. \method{} gains +5.4 to +10.4 points by iter-3, with the largest improvements on Deep Research and Scholarly QA. Best score in \textbf{bold}. }
  \label{tab:main_long}
\end{table}

\begin{table}[t]
  \centering
  \footnotesize
  \setlength{\tabcolsep}{3.2pt}
  \begin{tabular}{l ccc cccc c}
    \toprule
                               & \multicolumn{3}{c}{General QA} & \multicolumn{4}{c}{Multi-Hop QA} &                                                                                               \\
    \cmidrule(lr){2-4} \cmidrule(lr){5-8}
    Model                      & NQ                             & TriviaQA                         & PopQA         & HotpotQA      & 2Wiki         & MuSiQue       & Bamboogle     & Avg.          \\
    \midrule
    Qwen2.5-7B                 & 56.4                           & 71.6                             & 44.9          & 49.0          & 36.0          & 17.9          & 45.6          & 45.9          \\
    \quad GRPO$_{\text{data}}$ & 68.1                           & 77.0                             & 50.8          & 63.7          & \textbf{53.6} & 28.8          & 52.0          & 56.3          \\
    \quad \method\ iter1       & 64.2                           & 73.7                             & 47.2          & 57.8          & 45.8          & 26.6          & 51.2          & 52.4          \\
    \quad \method\ iter2       & 67.2                           & 76.8                             & 50.2          & \textbf{65.7} & 53.1          & 31.3          & 50.4          & 56.4          \\
    \quad \method\ iter3       & \textbf{74.4}                  & \textbf{81.0}                    & \textbf{53.5} & 65.6          & 52.3          & \textbf{32.6} & \textbf{58.4} & \textbf{59.7} \\
    \midrule
    Qwen3-8B                   & 62.8                           & 77.2                             & 48.7          & 56.8          & 52.3          & 24.4          & 56.0          & 54.0          \\
    \quad GRPO$_{\text{data}}$ & 69.0                           & 79.0                             & 52.5          & \textbf{71.5} & 57.6          & 32.7          & 62.3          & 60.7          \\
    \quad \method\ iter1       & 68.4                           & 78.9                             & 49.3          & 60.9          & 55.6          & 29.7          & 64.0          & 58.1          \\
    \quad \method\ iter2       & 70.5                           & \textbf{81.1}                    & 53.3          & 62.8          & 55.5          & 34.0          & 59.2          & 59.5          \\
    \quad \method\ iter3       & \textbf{71.9}                  & 80.9                             & \textbf{54.5} & 65.8          & \textbf{60.5} & \textbf{34.4} & \textbf{64.8} & \textbf{61.8} \\
    \midrule
    OLMo-3-7B                  & 53.1                           & 61.3                             & 45.4          & 41.7          & 32.4          & 15.7          & 30.4          & 40.0          \\
    \quad GRPO$_{\text{data}}$ & \textbf{73.0}                  & 64.0                             & \textbf{52.0} & \textbf{60.0} & 32.0          & 21.0          & \textbf{38.0} & 48.6          \\
    \quad \method\ iter1       & 57.6                           & 67.4                             & 46.0          & 46.2          & 39.5          & 16.7          & 27.2          & 42.9          \\
    \quad \method\ iter2       & 63.3                           & 71.3                             & 48.9          & 53.3          & \textbf{44.7} & 20.9          & 33.6          & 48.0          \\
    \quad \method\ iter3       & 67.3                           & \textbf{71.8}                    & 49.2          & 55.0          & 43.4          & \textbf{23.1} & 34.4          & \textbf{49.2} \\
    \bottomrule
  \end{tabular}
  \caption{\textbf{Short-form QA results.} \method{} trains exclusively on open-ended tasks, yet improves short-form QA by +7.8 to +13.8 points and surpasses GRPO$_{\text{data}}$ across models. This suggests \textbf{\method{} generalises more broadly than curated-prompt training.}}
  \label{tab:main_short}
\end{table}

\paragraph{Largest gains on research-intensive tasks.}
All three models see substantial gains that improve monotonically across iterations, inversely related to base model capacity, as \cref{tab:main_long,fig:avg_long_training} show: +10.4 for Qwen2.5-7B (24.4 $\to$ 34.8), +7.8 for OLMo-3-7B (30.7 $\to$ 38.5), and +5.4 for Qwen3-8B (37.7 $\to$ 43.1).
\method{}'s improvements are most pronounced on research-intensive tasks, with Deep Research and Scholarly QA gaining +11.1 points averaged across models, followed by planning (+8.4), creative writing (+4.3), and user assistance (+3.0)---consistent with \method{}'s training focus on document-grounded retrieval and synthesis.

\paragraph{\method{} matches curated-data training.}
Without curated prompts or frontier-model rubrics, \method{} matches GRPO$_{\text{data}}$ on average scores across open-ended benchmarks (\method{}/GRPO$_{\text{data}}$): 34.8/33.4 on Qwen2.5, 43.1/41.5 on Qwen3, and 38.5/39.0 on OLMo-3.
\method{} improves all models across task types, while GRPO$_{\text{data}}$ regresses on creative writing for Qwen3 ($-3.9$) and OLMo-3 ($-2.0$).
Because creative writing differs from the document-grounded training tasks, these gains suggest that corpus-grounded self-play builds synthesis skills that transfer across task types (\cref{sec:analysis:decomp}).
\method{} is comparable even on deep research, the setting most aligned with GRPO$_{\text{data}}$'s training data.

\looseness-1
\paragraph{Short-form QA transfer.}
\method{} trains exclusively on open-ended tasks, yet its short-form QA averages reach 59.7 on Qwen2.5, 49.2 on OLMo-3, and 61.8 on Qwen3---gains of 7.8--13.8 points over the respective base models, as \cref{tab:main_short} shows.
Across three models, \method{} yields comparable gains on general QA (+9.2) and multi-hop QA (+11.0), with a modest advantage on the latter, suggesting that training on open-ended tasks builds retrieval and reasoning skills that generalise to short-form QA.
On all three backbones, \method{} surpasses GRPO$_{\text{data}}$ on the short-form average by as much as 3.4 points.
This suggests \method{} yields broader capability gains than training on curated prompts.

\begin{wrapfigure}[18]{r}{0.38\textwidth}
  \centering
  \vspace{-1.5em}
  \includegraphics[width=0.37\textwidth]{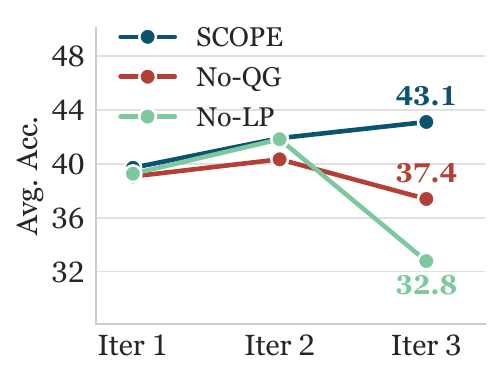}
  \vspace{-1em}
  \caption{ Reward design ablation on Qwen3-8B. \textbf{No-QG} removes the Challenger's quality gates; \textbf{No-LP} removes the Solver's cosine length penalty. Reward hacking compounds over iterations and both mitigations are necessary for sustained improvement.}
  \label{fig:reward_hacking}
  \vspace{-1em}
\end{wrapfigure}

\paragraph{Reward design ablation.}
To confirm that the quality gates and cosine length penalty each prevent a distinct failure mode, we ablate them individually on Qwen3-8B, as shown in \cref{fig:reward_hacking}.
All three conditions start from comparable iter-1 performance and track closely through iter-2, but the ablated variants diverge sharply at iter-3: No-QG drops to 37.4 and No-LP collapses to 32.8---both below the 37.7 base model---while \method{} continues improving to 43.1.
The two ablations expose distinct failure modes.
Without quality gates, the Challenger drifts towards generic tasks that ignore the source document but still elicit moderate Solver rubric scores, maximising the difficulty reward while degrading the training signal as tasks converge to a narrow distribution.
Without the length penalty, the Solver exploits the rubric judge's tendency to score longer outputs more favourably; on Qwen3-8B, this escalates to compressing thinking tokens to allocate more of the fixed context window to the visible response, eventually collapsing training entirely.

\section{Analysis}
\label{sec:analysis}

We study \method{} on Qwen3-8B, the strongest base model in our experiments, to understand what drives its improvement and where it plateaus.
Since \method{} depends on all three roles working in concert, we examine a single question from three angles:
(1)~whether the Challenger must co-evolve with the Solver,
(2)~what the Solver learns, and
(3)~what makes the Self-Judge effective.

\begin{figure}[t]
  \centering
  \includegraphics[width=.75\textwidth]{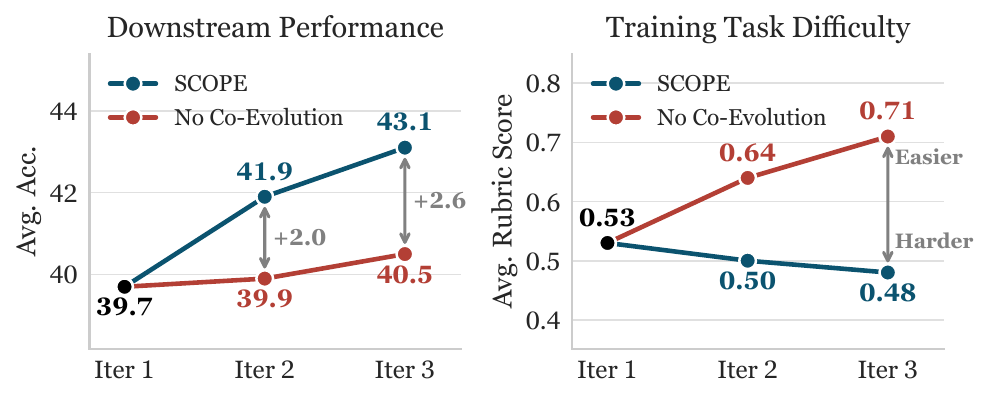}
  \caption{\textbf{The Challenger must co-evolve with the Solver.} (a)~On downstream benchmarks, \method{} improves by +3.4 points over three iterations, whereas a frozen Challenger limits gains to +0.8. (b)~Co-evolution keeps average rubric score near the 0.5 sweet spot, while a frozen Challenger drifts to trivially easy tasks (rubric score 0.71 by iter\,3), weakening the training signal.}
  \label{fig:frozen_ablation}
\end{figure}

\paragraph{Co-evolution is necessary.}
\label{sec:analysis:frozen}
We ask whether the Challenger must co-evolve with the Solver, or whether a fixed Challenger suffices.
We compare \method{} (Challenger and Solver both evolve across three iterations) against a variant that fixes the Challenger at its iter-1 checkpoint while the Solver continues training on fresh rollouts from this static Challenger (``No Co-Evolution'' in \cref{fig:frozen_ablation}).
Both conditions share the iter-1 checkpoint (step~20).

\cref{fig:frozen_ablation} reveals a widening gap between the two conditions.
From iter-1 to iter-3, \method{} gains +3.4 points on average across open-ended benchmarks (39.7 $\to$ 43.1), while the frozen Challenger condition gains only +0.8 (39.7 $\to$ 40.5).
The gap widens over iterations---2.0 points at iter-2, 2.6 at iter-3---because without co-evolution the Challenger's tasks no longer challenge the improving Solver, diminishing the training signal.
Training-time rubric scores in \cref{fig:frozen_ablation} (right) show the same pattern: under co-evolution, mean scores remain near the optimal difficulty of 0.5 across iterations (0.53$\to$0.50$\to$0.48), indicating tasks stay at the Solver's frontier; without Challenger evolution, scores rise to 0.71 by iter-3, reflecting tasks the Solver has outgrown.
Sustained self-play improvement on open-ended tasks therefore requires both policies to co-evolve.

\begin{wrapfigure}[19]{r}{0.35\textwidth}
  \centering
  \vspace{-1.5em}
  \includegraphics[width=0.35\textwidth]{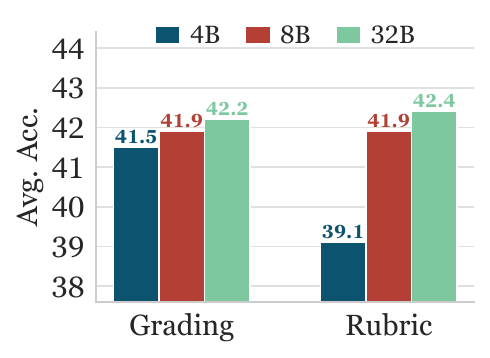}
  \vspace{-1em}
  \caption{\textbf{Self-Judge effectiveness depends more on rubrics than grading.} We vary the model size of either the grader or rubric generator while holding the other fixed at 8B. Average open-ended benchmark performance is stable across grader sizes, but drops sharply with a 4B rubric generator.}
  \label{fig:judge_sensitivity}
  \vspace{-1em}
\end{wrapfigure}

\paragraph{Both retrieval and synthesis improve.}
\label{sec:analysis:decomp}

As \cref{tab:main_long,tab:main_short} show, \method{}'s gains span both open-ended and short-form tasks, raising the question of whether training improves the Solver's ability to retrieve relevant information or to consolidate it into a coherent answer.
We disentangle these components for Qwen3-8B via a controlled replay experiment.\footnote{We compare iter-1 and iter-3 rather than base and iter-3 because iter-1 is the earliest checkpoint that has learned the multi-turn search format; the comparison would otherwise confound format learning with genuine improvements in retrieval and synthesis.}
Starting from the iter-1 Solver as baseline, we swap in one component at a time: the iter-3 answer generator or the iter-3 search trajectory.

\begin{table}[t]
  \centering
  \small
  \begin{tabular}{@{}ll|r@{\,}l r@{\,}l r@{\,}l r@{\,}l@{}}
    \toprule
    Search & Answer & \multicolumn{2}{c}{NQ} & \multicolumn{2}{c}{HotpotQA}         & \multicolumn{2}{c}{ResearchQA} & \multicolumn{2}{c}{HealthBench}                                                                                                  \\
    \midrule
    Iter-1 & Iter-1 & 68.4                   &                                      & 60.9                           &                                      & 47.8 &                                      & 26.2 &                                      \\
    Iter-1 & Iter-3 & 71.1                   & {\color{green!50!black}\small(+2.7)} & 61.8                           & {\color{green!50!black}\small(+0.9)} & 51.3 & {\color{green!50!black}\small(+3.5)} & 27.5 & {\color{green!50!black}\small(+1.3)} \\
    Iter-3 & Iter-1 & 69.0                   & {\color{green!50!black}\small(+0.6)} & 64.5                           & {\color{green!50!black}\small(+3.6)} & 50.0 & {\color{green!50!black}\small(+2.2)} & 26.6 & {\color{green!50!black}\small(+0.4)} \\
    \bottomrule
  \end{tabular}
  \caption{\textbf{\method{} improves retrieval and synthesis.}
    Starting from the Iter-1 Solver, each row replaces one component with its Iter-3 counterpart.
    Replacing only the answer generator (Iter-1 search fixed) measures synthesis: turning fixed evidence into the final answer.
    Replacing only the search trajectory (Iter-1 generator fixed) measures retrieval: finding task-relevant evidence.
    Gains appear across all benchmarks, with retrieval larger on multi-hop tasks and synthesis larger on single-hop and knowledge-mismatched tasks.}
  \label{tab:decomp}
\end{table}

\cref{tab:decomp} shows that \method{} improves both retrieval and synthesis across all four benchmarks, with the dominant source of gain tracking each task's bottleneck.
When retrieval is the limiting factor---as in multi-hop \emph{HotpotQA}, which requires chaining queries across bridging entities---retrieval contributes the larger gain (+3.6 vs.\ +0.9 from synthesis).
Conversely, when relevant evidence is accessible through simple queries or the knowledge source is less aligned with the task, synthesis contributes more: +2.7 vs.\ +0.6 on single-hop \emph{NQ}, and +1.3 vs.\ +0.4 on \emph{HealthBench}.
\emph{ResearchQA}, which demands both multi-step retrieval and long-form integration, shows sizable gains from both (+2.2 and +3.5).
These gains confirm that \method{} improves both retrieval and synthesis across task types.

\paragraph{Rubric quality matters more than grading.}
\label{sec:analysis:judge}

\method{} uses a frozen copy of the base model as Judge, requiring no external supervision.
To understand what makes this viable, we separately scale the Judge's two roles (rubric generation and grading) from 4B to 32B, holding the other fixed at 8B (\cref{fig:judge_sensitivity}).
All conditions branch from a shared iter-1 checkpoint and run iter-2 Solver training only, after the Solver has acquired the multi-turn search format; this isolates judge quality from format learning.
Varying only the grader changes the open-ended benchmark average by just 0.7 points (41.5--42.2), whereas a 4B rubric generator drops the average by 2.8 points.
Scaling rubric generation to 32B adds only +0.5 over 8B, suggesting that rubric specificity, not grading capacity, is the binding factor.

To explain this gap, we examine the generated rubrics (\cref{tab:rubric_quality_example} in \cref{app:reward} shows a representative case).
We find that 4B rubrics often omit document-specific facts, producing criteria that any on-topic response can satisfy.
In contrast, 8B and 32B rubrics tend to ground their criteria in concrete source details---dates, dollar amounts, and named entities---and produce largely overlapping requirements; the few cases where 32B differs tend to be analytical, such as testing the reasoning behind an event rather than just its occurrence.

\section{Related Work}
\label{sec:related}

\paragraph{Data-free self-play.}
Self-play achieved superhuman performance in various games~\citep{silver2016mastering,silver2017mastering,vinyals2019grandmaster,brown2019superhuman,openai2019dota,meta2022diplomacy},
demonstrating that agents can discover complex strategies without human
data~\citep{baker2019emergent}.
Recent work extends self-play to LLM post-training: Absolute
Zero~\citep{azr} and R-Zero~\citep{rzero} co-evolve task generation and
solving via GRPO but reward code execution or math answer correctness;
SPICE~\citep{spice} and Dr.~Zero~\citep{drzero} ground self-play in an
unlabelled corpus but still require verifiable short answers;
EVA~\citep{eva} and OpenSIR~\citep{kwan2025opensir} broaden coverage but
rely on frontier reward models or remain in verifiable domains.
All prior data-free self-play depends on deterministic correctness or external
reward models; \method{} preserves co-evolution while replacing answer
correctness with rubric-based self-judging.

\paragraph{Rubric rewards for open-ended tasks.}
Rubric-based RL replaces scalar rewards with task-specific evaluation criteria.
Prompt-based approaches generate rubrics via off-the-shelf
models~\citep{rar,openrs} or update them on-policy with search-grounded
knowledge~\citep{drtulu}, while \citet{meta2025rpg} extract goals and rubrics
from conference papers via frontier models.
Complementary work trains reward models to reason through rubric
criteria~\citep{rmr1}, co-trains rubric generators and policies with
correctness labels~\citep{rlcer}, learns rubric generators or judges from
preference data~\citep{openrubrics,rubricarm,queryrubrics}, or formalises rubric
rewards as weighted scores with online criterion
elicitation~\citep{scaleai2025onlinerubrics}.
\method{} is the first rubric-based self-play method that requires no curated
prompts, no external labels, and no frontier supervision.

\section{Conclusion}
\label{sec:conclusion}

We presented \method{}, which extends data-free self-play to open-ended tasks
through document-grounded rubric decomposition: a frozen self-judge writes
task-specific criteria from source documents, providing the reward signal
that prior methods could only obtain from verifiable answers.
Across three 7--8B model families, we improve open-ended performance by +5.4
to +10.4 points and match or exceed GRPO$_{\text{data}}$ trained on
${\sim}$9K curated prompts, without any curated data or frontier-model
supervision.
These gains transfer beyond the training domain to short-form QA and
creative writing, despite training solely on information-dense open-ended tasks.
Our analysis shows that co-evolution is necessary, as a frozen Challenger
fails to propose tasks that offer learning value.
\method{} strengthens both retrieval and synthesis, with the relative
contribution varying by task.
We further find that rubric generation quality, not grading capacity, is the bottleneck for self-judging: once criteria are specific and document-grounded, scaling either component yields diminishing returns.
Together, these findings suggest that \method{} represents an important first step towards scalable, data-free self-improvement on open-ended tasks.


\bibliographystyle{plainnat}
\bibliography{references}

\appendix


\section{Limitations}
\label{app:limitations}

The multi-stage pipeline (Challenger training, rubric generation, and task filtering) requires more compute than single-stage GRPO on curated prompts.
This overhead limited our experiments to 7--8B models; whether the gains hold at larger scales (e.g., 32B) remains an open question.
Self-play targets the post-data regime where curated supervision has been exhausted, so trading additional compute for continued improvement may be justified.

\paragraph{Broader impacts.}
\method{} may reduce dependence on human-curated prompts and frontier-model judges, making self-improvement more accessible to open-weight models and research groups that lack proprietary supervision pipelines.
A risk specific to data-free self-play is that the Challenger may generate tasks involving sensitive, biased, or inappropriate content; for example, tasks that synthesise medical misinformation or reproduce stereotypes from the source corpus.
Because the Solver trains on these tasks, harmful content can propagate across iterations and entrench undesirable behaviours.
The quality gates (\cref{sec:method:sustainability}) mitigate this in part by filtering ill-formed or source-irrelevant tasks, but they do not screen for harmful content explicitly.
Future deployments should add content-safety filtering at the task-generation stage and monitor the training distribution for harmful topic drift.


\section{Additional Results}
\label{app:additional_results}

\subsection{Training Dynamics}
\label{app:training_dynamics}

\begin{figure}[htbp]
  \centering
  \includegraphics[width=\textwidth]{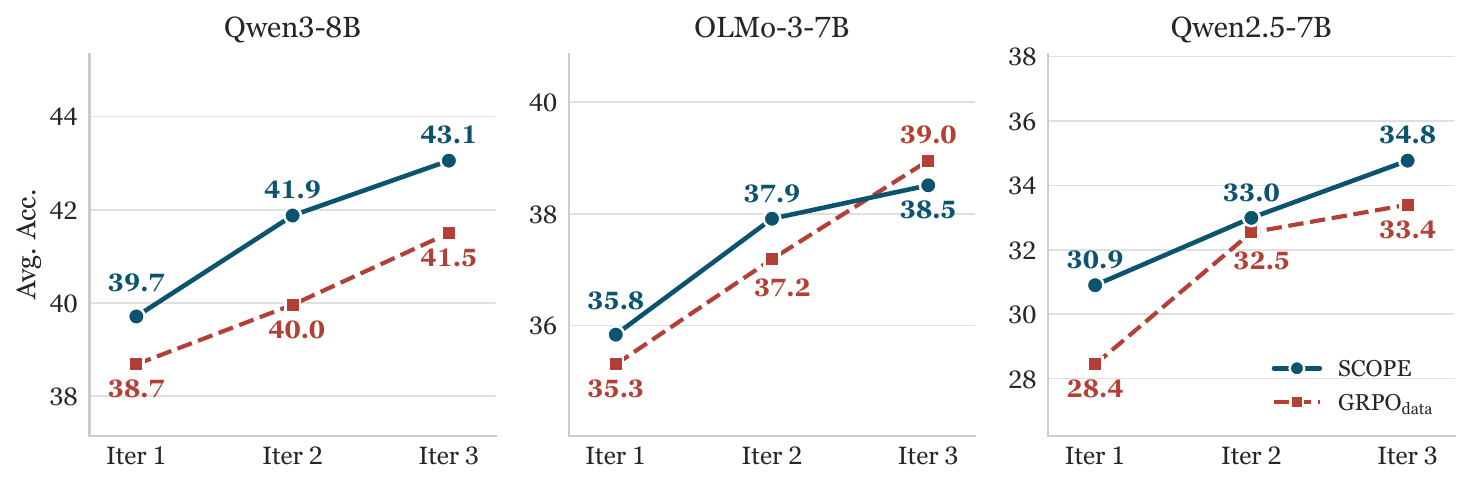}
  \caption{\textbf{Average open-ended benchmark score across three training iterations.} \method{} matches or exceeds GRPO$_{\text{data}}$ at nearly every iteration, with only OLMo-3-7B at iteration~3 reversing this pattern. Both methods show diminishing gains, and weaker base models benefit most from training.}
  \label{fig:avg_long_training}
\end{figure}
\paragraph{Performance over iterations.}
\cref{fig:avg_long_training} tracks average open-ended benchmark scores for \method{} and GRPO$_{\text{data}}$ across three iterations.
Despite using no curated prompts, \method{} matches or exceeds GRPO$_{\text{data}}$ at every iteration across all three model families, with the sole exception of OLMo-3-7B at iteration~3 where GRPO$_{\text{data}}$ edges ahead by 0.5 points.
Per-iteration gains diminish for both methods, and weaker base models benefit most (+10.4 for Qwen2.5-7B vs.\ +5.4 for Qwen3-8B).

\begin{figure}[htbp]
  \centering
  \includegraphics[width=\textwidth]{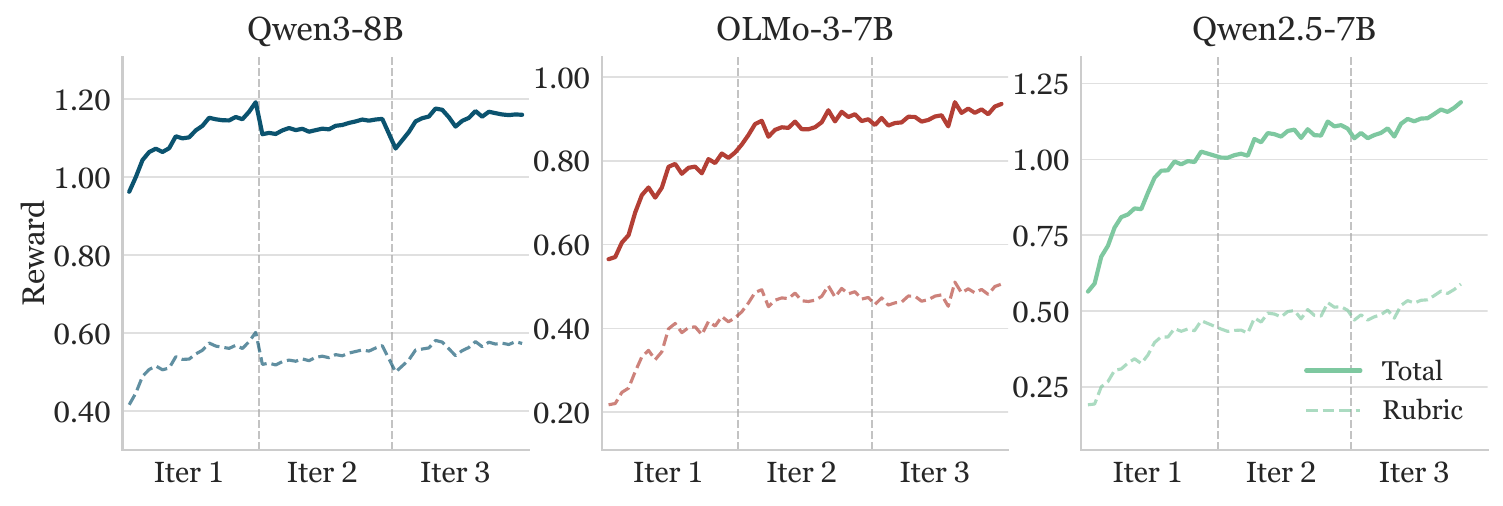}
  \caption{\textbf{Training-time reward across Solver iterations} (total: solid; rubric: dashed).
    Rubric reward dips at iteration boundaries as the co-evolved Challenger generates harder tasks, then recovers as the Solver adapts.
    Total reward rises more steadily, driven by stable format and search components.}
  \label{fig:training_acc_reward}
\end{figure}

\paragraph{Reward evolution.}
\cref{fig:training_acc_reward} shows both total and rubric reward during Solver training; both increase steadily across all three models, confirming that GRPO training consistently improves the Solver.
Within this overall trend, rubric reward dips at iteration boundaries for Qwen3-8B and OLMo-3-7B: once the Solver has improved on the current tasks, the co-evolved Challenger responds with harder tasks calibrated to elicit an average Solver score near $\tau{=}0.5$.
Qwen2.5-7B does not exhibit these dips because its rubric reward remains below 0.5 throughout, so the difficulty increase between iterations is less pronounced.
The total reward climbs more steadily because format and search rewards are less affected by task difficulty.

\begin{figure}[htbp]
  \centering
  \includegraphics[width=\textwidth]{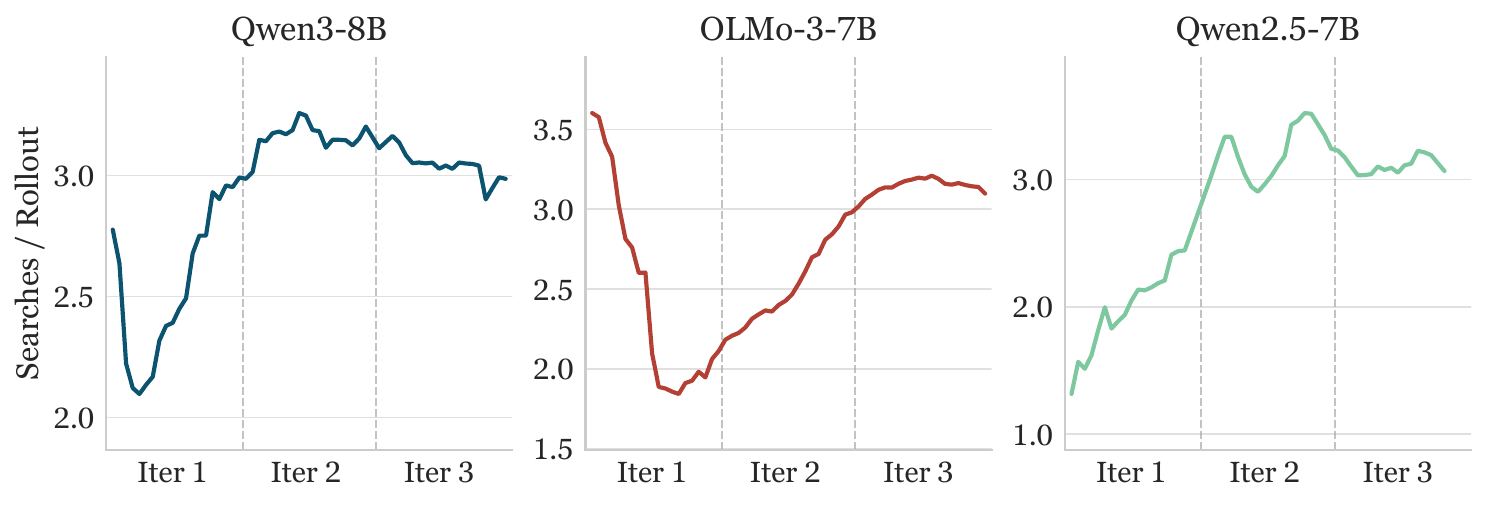}
  \caption{\textbf{Mean valid search calls per rollout during Solver training.}
    All three models reduce search usage early in iteration~1, then recover and converge near three searches per rollout by iteration~3, matching the search reward saturation point.}
  \label{fig:training_search_count}
\end{figure}

\paragraph{Search behaviour.}
\cref{fig:training_search_count} reveals a shared drop-then-recovery pattern: all three Solvers reduce search calls early in iteration~1, then increase usage through iterations~2--3, converging near three searches per rollout---the search reward saturation point.
We inspected the rollouts to understand this pattern. Early in training, the Solver relies on parametric knowledge and issues few searches, producing the initial dip even as rubric reward rises; in later iterations, it learns that retrieval improves answer quality and increases its search usage, despite the small search bonus.
OLMo-3-7B shows the sharpest contraction (${\sim}3.5$ to ${\sim}1.9$) before steadily climbing back.

\subsection{Extended Training Iterations}
\label{app:extended_training}

\begin{table}[htbp]
  \centering
  \footnotesize
  \setlength{\tabcolsep}{3.2pt}
  \begin{tabular}{l cc cc c cc c c}
    \toprule
                    & \multicolumn{2}{c}{Deep Research} & \multicolumn{2}{c}{Scholarly QA} & Planning      & \multicolumn{2}{c}{User Assist.} & Creative      &                                                               \\
    \cmidrule(lr){2-3} \cmidrule(lr){4-5} \cmidrule(lr){6-6} \cmidrule(lr){7-8} \cmidrule(lr){9-9}
    Iteration       & DRB                               & Rubrics                          & ResQA         & SQAv2                            & ResPlan       & HealthB.      & WildB.        & AH-CW         & Avg.          \\
    \midrule
    Base model      & 49.2                              & 24.2                             & 40.0          & 37.5                             & 53.9          & 25.9          & 45.7          & 24.9          & 37.7          \\
    \method\ iter 1 & 51.3                              & 25.8                             & 47.8          & 38.5                             & 55.3          & 26.2          & 46.2          & 26.5          & 39.7          \\
    \method\ iter 2 & 55.5                              & 30.1                             & 51.6          & 39.8                             & 56.8          & 26.8          & 46.9          & 27.6          & 41.9          \\
    \method\ iter 3 & 57.6                              & 31.3                             & 53.7          & 41.8                             & 57.7          & 28.1          & 46.1          & 28.4          & 43.1          \\
    \method\ iter 4 & 58.6                              & 32.1                             & 55.0          & 42.5                             & 58.4          & 27.8          & 46.9          & 29.5          & 43.9          \\
    \method\ iter 5 & 59.1                              & 32.8                             & 55.7          & 43.0                             & 58.5          & 28.6          & 47.0          & \textbf{30.5} & 44.4          \\
    \method\ iter 6 & \textbf{59.6}                     & \textbf{33.4}                    & \textbf{56.8} & \textbf{43.4}                    & \textbf{58.8} & \textbf{28.8} & \textbf{47.3} & 30.3          & \textbf{44.8} \\
    \bottomrule
  \end{tabular}
  \caption{\textbf{Extended training beyond three iterations} on Qwen3-8B with identical hyperparameters. Average performance rises monotonically from $37.7$ to $44.8$, with diminishing per-iteration gains ($+2.0$ to $+0.4$). Iterations 1--3 account for +5.4 of the total +7.1 point gain. Best per benchmark in \textbf{bold}.}
  \label{tab:extended_long}
\end{table}

\begin{figure}[htbp]
  \centering
  \includegraphics[width=0.62\textwidth]{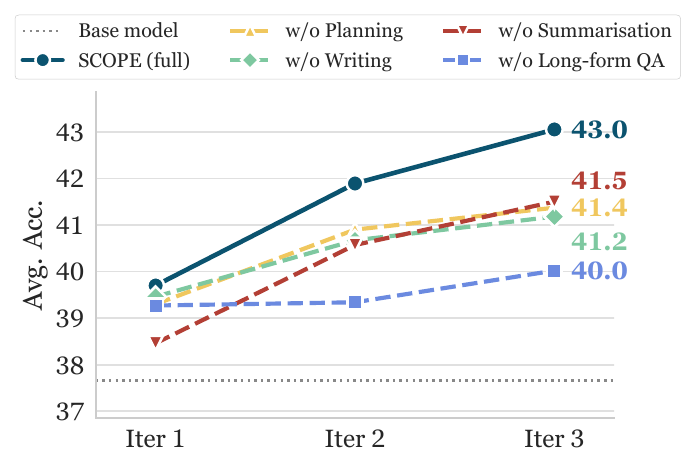}
  \caption{\textbf{Leave-one-out domain ablation across iterations} on Qwen3-8B, averaged over open-ended benchmarks.
    Dashed grey line: base model.
    All ablations improve monotonically, but the full four-domain mixture improves fastest; its margin over the best leave-one-out variant widens from $0.24$ (iter~1) to $0.99$ (iter~2) to $1.55$ (iter~3), indicating that cross-domain transfer compounds with training.}
  \label{fig:loo_iterations}
\end{figure}

A natural question is whether \method{} continues to improve beyond three iterations or eventually saturates.
We extend training to six iterations on Qwen3-8B, the strongest model in our comparison, keeping all hyperparameters identical.
As \cref{tab:extended_long} shows, per-iteration gains shrink steadily but remain positive through iteration~6, and three iterations yield the bulk of improvement ($+5.4$ of $+7.1$).
The sustained gains through iteration~6 demonstrate that Challenger co-evolution enables sustainable self-improvement.

\subsection{Leave-One-Out Domain Ablation}
\label{app:loo_domain_ablation}

We train four leave-one-out variants of \method{} on Qwen3-8B, each excluding one Challenger task domain (long-form QA, summarisation, planning, or writing) while keeping all other settings fixed, to isolate each domain's contribution to open-ended performance.

\paragraph{Iteration-level trends.}
Every leave-one-out variant improves monotonically across iterations (\cref{fig:loo_iterations}), confirming that \method{} training is robust to the composition of the task mixture.
However, the full four-domain mixture improves fastest: its margin over the strongest leave-one-out variant compounds from $0.24$ at iteration~1 to $0.99$ at iteration~2 and $1.55$ at iteration~3.
This widening gap suggests that each domain cultivates distinct capabilities whose benefits compound when combined.

\begin{table}[htbp]
  \centering
  \footnotesize
  \setlength{\tabcolsep}{3.2pt}
  \begin{tabular}{l cc cc c cc c c}
    \toprule
                            & \multicolumn{2}{c}{Deep Research} & \multicolumn{2}{c}{Scholarly QA} & Planning      & \multicolumn{2}{c}{User Assist.} & Creative      &                                                               \\
    \cmidrule(lr){2-3} \cmidrule(lr){4-5} \cmidrule(lr){6-6} \cmidrule(lr){7-8} \cmidrule(lr){9-9}
    Config                  & DRB                               & Rubrics                          & ResQA         & SQAv2                            & ResPlan       & HealthB.      & WildB.        & AH-CW         & Avg.          \\
    \midrule
    \method{}               & \textbf{57.6}                     & \textbf{31.3}                    & 53.7          & \textbf{41.8}                    & 57.7          & 28.1          & \textbf{46.1} & \textbf{28.4} & \textbf{43.1} \\
    \quad w/o Planning      & 55.3                              & 29.4                             & 51.5          & 38.5                             & 54.8          & \textbf{30.9} & 45.1          & 25.5          & 41.4          \\
    \quad w/o Writing       & 55.7                              & 29.2                             & \textbf{54.7} & 39.0                             & 58.1          & 30.2          & 42.4          & 20.3          & 41.2          \\
    \quad w/o Summarisation & 54.9                              & 29.9                             & 51.4          & 38.3                             & \textbf{58.1} & 28.9          & 44.2          & 26.3          & 41.5          \\
    \quad w/o Long-form QA  & 53.9                              & 27.4                             & 49.2          & 34.8                             & 57.0          & 30.5          & 43.2          & 24.3          & 40.0          \\
    \bottomrule
  \end{tabular}
  \caption{\textbf{Per-benchmark iter-3 results for leave-one-out domain ablation} on Qwen3-8B.
    Long-form QA is the most broadly impactful domain ($-3.0$ avg, worst on 4/8 benchmarks); writing and planning show concentrated effects on closely related benchmarks (Arena-Hard $-8.1$, ResearchPlanGen $-2.9$). Best per benchmark in \textbf{bold}.}
  \label{tab:loo_benchmarks}
\end{table}

\paragraph{Per-benchmark breakdown.}
\cref{tab:loo_benchmarks} shows a clear hierarchy at iteration~3.
Long-form QA is the most foundational: removing it causes the largest average drop ($-3.0$~points, roughly twice that of any other single-domain ablation) and the worst scores on four of eight benchmarks (SQA-CS-V2 $-7.0$, ResearchQA $-4.5$, ResearchRubrics $-3.9$, DRB-RACE $-3.7$), all centred on retrieval, reasoning, and synthesis.
The remaining three domains cluster in average impact ($-1.5$ to $-1.9$) yet specialise distinctly: removing writing produces the steepest drops on creative and user-facing benchmarks (Arena-Hard $-8.1$, WildBench $-3.7$), removing planning primarily degrades structured reasoning (ResearchPlanGen $-2.9$), and removing summarisation spreads a diffuse deficit across benchmarks without dominating any single one.
Each domain thus provides a distinct training signal; their combination compounds rather than plateaus.

\subsection{Reference Comparison with Related Methods}
\label{app:contemporary_comparison}

\begin{table}[htbp]
  \centering
  \footnotesize
  \setlength{\tabcolsep}{4pt}
  \begin{tabular}{l cccc}
    \toprule
                     & \method{} (ours) & DR~Tulu      & WebExplorer & WebThinker-R1 \\
    \midrule
    \makecell{Base                                                                   \\Model}           & Qwen3-8B         & Qwen3-8B              & Qwen3-8B      & QwQ-32B       \\ \midrule
    \makecell{Task                                                                   \\Type}            & Open-ended       & Deep research         & Short-form QA & Short-form QA \\ \midrule
    \makecell{\# Seed                                                                \\Prompts}      & 0                & ${\sim}$9K            & ${\sim}$12K   & ${\sim}$3K    \\ \midrule
    \makecell{Rubric                                                                 \\Source}         & Self-generated   & GPT-4.1               & ---           & ---           \\ \midrule
    \makecell{Judge} & Self-judge       & GPT-4.1-mini & ---         & ---           \\ \midrule
    \makecell{Training                                                               \\Recipe}            & GRPO (60 steps)  & \makecell{SFT $\rightarrow$ GRPO\\(1.9K steps)} & \makecell{SFT $\rightarrow$ GRPO\\(${\sim}$380 steps)} & DPO (2 iters) \\ \midrule
    \makecell{External                                                               \\Model\\Dependency} & None             & \makecell{GPT-4.1\\GPT-4.1-mini\\GPT-5} & \makecell{DeepSeek-V3\\Gemini 2.5 Flash} & Qwen2.5-72B-Instruct \\
    \bottomrule
  \end{tabular}
  \caption{\textbf{Training setup comparison with related methods.} \method{} is the only method that self-generates its entire training curriculum with no frontier model dependencies. WebExplorer and WebThinker train on verifiable QA tasks, where rubric-based evaluation is unnecessary (---).}
  \label{tab:contemporary_setup}
\end{table}

\begin{table}[htbp]
  \centering
  \footnotesize
  \setlength{\tabcolsep}{3.2pt}
  \begin{tabular}{l cc cc c cc c c}
    \toprule
                     & \multicolumn{2}{c}{Deep Research} & \multicolumn{2}{c}{Scholarly QA} & Planning      & \multicolumn{2}{c}{User Assist.} & Creative      &                                                               \\
    \cmidrule(lr){2-3} \cmidrule(lr){4-5} \cmidrule(lr){6-6} \cmidrule(lr){7-8} \cmidrule(lr){9-9}
    Method           & DRB                               & Rubrics                          & ResQA         & SQAv2                            & ResPlan       & HealthB.      & WildB.        & AH-CW         & Avg.          \\
    \midrule
    \method{} (ours) & 57.6                              & 31.3                             & 53.7          & \textbf{41.8}                    & 57.7          & \textbf{28.1} & \textbf{46.1} & \textbf{28.4} & 43.1          \\
    DR~Tulu          & \textbf{70.1}                     & \textbf{40.0}                    & \textbf{67.5} & 41.0                             & \textbf{68.4} & 25.3          & 36.0          & 4.4           & \textbf{44.1} \\
    WebExplorer      & 55.6                              & 29.2                             & 55.4          & 39.0                             & 50.3          & 26.4          & 41.7          & 14.6          & 39.0          \\
    WebThinker-R1    & 40.2                              & 9.5                              & 47.0          & 37.8                             & 23.0          & 12.5          & 36.5          & 5.7           & 26.5          \\
    \bottomrule
  \end{tabular}
  \caption{\textbf{Per-benchmark comparison with related methods} on open-ended benchmarks. \method{} is the most balanced, achieving the best score on 4/8 benchmarks (SQAv2, HealthBench, WildBench, Arena-Hard). DR~Tulu leads on deep research tasks but scores below the base model on Arena-Hard (4.4) and WildBench (36.0). Best per benchmark in \textbf{bold}.}
  \label{tab:contemporary_benchmarks}
\end{table}

\paragraph{Compared methods.}
Several recent methods also train language models for open-ended tasks with search capabilities~\citep{drtulu, liu2025webexplorer, li2025webthinker}.
We consider only methods with publicly released checkpoints, enabling re-evaluation under our protocol.
Critically, all of them rely on curated or externally sourced prompts (3K--12K), whereas \method{} generates its entire training curriculum from a raw corpus.
They also differ from \method{} along additional axes: DR~Tulu and WebExplorer require SFT warmup on frontier-model trajectories and frontier-model judges, and all three provide richer tool access (browsing, web crawling) than \method{}'s search-only setup.
Because these overlapping confounds prevent controlled comparison, we report results as a reference evaluation rather than including them in the main results.

\paragraph{Evaluation protocol.}
To ensure a uniform evaluation, we replace each method's native tools (browsing, paper search, web crawling) with \method{}'s single retrieval endpoint and grade all outputs with the same LLM judge, while preserving each method's original prompt format and tool-calling conventions.
\cref{tab:contemporary_setup} summarises the training differences; \cref{tab:contemporary_benchmarks} reports per-benchmark scores.

\paragraph{Results.}
Despite using no curated prompts and no frontier model supervision, \method{} achieves 43.1 on average, within 1.0 point of DR~Tulu (44.1) and ahead of WebExplorer (39.0) and WebThinker-R1 (26.5), while scoring highest on four of eight benchmarks.
The gap is concentrated on research-heavy benchmarks, where DR~Tulu's curated prompts and frontier-model rubrics closely match the evaluation distribution.
On user-facing and creative tasks, \method{} leads despite requiring no external data or frontier-model supervision.


\section{Detailed Training Algorithm}
\label{app:algorithm}

\cref{alg:scope} gives the complete pseudocode of the \method{} training loop, including rollout collection, quality gating, rubric generation, and difficulty filtering.

\begin{algorithm}[ht]
  \caption{\method{} training loop.}
  \label{alg:scope}
  \begin{algorithmic}[1]
    \Require Corpus $\corpus$, base model $M_0$, iterations $T$, batch size $B$, difficulty window $[\ell, u]$
    \State Challenger $C_0 \gets M_0$, Solver $S_0 \gets M_0$, Judge $J \gets M_0$
    \For{$t = 1, \ldots, T$}
    \Statex \hspace{\algorithmicindent}\textbf{--- Stage 1: Train Challenger ---}
    \For{$b \gets 1$ to $B$}
    \State Sample document $d \sim \corpus$
    \State Generate task $(o^C,\, q) \gets \textsc{Rollout}(C_{t-1},\, d)$
    \If{$q$ passes quality gates $\mathrm{QG}(q, \mathbf{d})$} \Comment{Estimate task difficulty}
    \State Generate rubric $\CJ \gets \textsc{GenRubric}(q, \mathbf{d})$
    \State Collect $N$ solver rollouts $o^S_{1:N} \gets \textsc{Rollout}(S_{t-1},\, q,\, N)$
    \State Grade rollouts $\bar{g} \gets \textsc{Grade}(o^S_{1:N},\; \CJ)$
    \State Compute reward $R_C$ \Comment{\cref{eq:challenger_reward}}
    \Else
    \State $R_C \gets r_{\mathrm{fmt}}(o^C)$ \Comment{Format reward only}
    \EndIf
    \EndFor
    \State Update $C_{t-1} \to C_t$ via $\textsc{GRPO}$
    \Statex \hspace{\algorithmicindent}\textbf{--- Stage 2: Train Solver ---}
    \State $\mathcal{T} \gets \emptyset$ \Comment{Collect difficulty-filtered tasks}
    \For{each document $d \in \textsc{Sample}(\corpus)$}
    \State Generate task $(\_, q) \gets \textsc{Rollout}(C_t,\, d)$
    \State Generate rubric $\CJ \gets \textsc{GenRubric}(q, \mathbf{d})$
    \State Grade $N$ solver rollouts: $\bar{g} \gets \textsc{Grade}(S_{t-1},\, q,\, \CJ,\, N)$
    \If{$\mathrm{QG}(q, \mathbf{d}) \;\land\; \ell \leq \bar{g} \leq u$}
    \State $\mathcal{T} \gets \mathcal{T} \cup \{(q,\, \CJ)\}$
    \EndIf
    \EndFor
    \For{$b \gets 1$ to $B$}
    \State Sample task $(q, \CJ) \sim \mathcal{T}$
    \State Generate response $o^S \gets \textsc{Rollout}(S_{t-1},\, q)$
    \State Compute reward $R_S$ \Comment{\cref{eq:solver_reward}}
    \EndFor
    \State Update $S_{t-1} \to S_t$ via $\textsc{GRPO}$
    \EndFor
    \State \Return $C_T,\, S_T$
  \end{algorithmic}
\end{algorithm}

\section{Theoretical Justification of the Difficulty Reward}
\label{app:diff}

The difficulty reward $f_{\mathrm{diff}}(\bar{g};\,\tau)$ (\cref{eq:diff}) peaks at mean rubric score $\bar{g} = \tau = \tfrac{1}{2}$, targeting the regime of maximum feedback variance.
Under GRPO (\cref{eq:advantage}), group-relative advantages $A_i = (R_i - \operatorname{mean}(R_{1:G})) / \operatorname{std}(R_{1:G})$ drive policy updates.
With population standard deviation, $\sum_i A_i^2 = G$ exactly, so the advantage \emph{magnitude} is constant; the mechanism by which extreme difficulty degrades training is the breakdown of the \emph{empirical} standard deviation $\hat\sigma_G$ in the denominator.
For rubric scores $g \in [0,1]$ with population mean $\mu_q = \mathbb{E}[g \mid q]$, the Bhatia--Davis inequality~\citep{Bhatia01042000} gives
\begin{equation}
  \operatorname{Var}(g \mid q) \;\leq\; \mu_q(1 - \mu_q) \;\leq\; \tfrac{1}{4},
  \label{eq:app_std}
\end{equation}
with equality for binary outcomes and the right bound tight at $\mu_q = \tfrac{1}{2}$.
As $\mu_q \to 0$ or $1$, $\operatorname{Var}(g \mid q) \to 0$, so $\hat\sigma_G \to 0$ and $A_i$ becomes either undefined or dominated by sampling noise.
The difficulty reward $f_{\mathrm{diff}}$ and the variance bound $\mu_q(1{-}\mu_q)$ share the same unique maximiser at $\mu_q = \tfrac{1}{2}$ and are both monotone decreasing in $|\mu_q - \tfrac{1}{2}|$, so $f_{\mathrm{diff}}$ ranks tasks by feedback variance in the same order as the bound.
The difficulty filter $\bar{g} \in [\ell, u]$ with $\ell = 0.2$, $u = 0.8$ complements the reward by lower-bounding $\mu_q(1{-}\mu_q) \geq \ell(1{-}\ell) = 0.16$, preventing the variance collapse that stalls training at the extremes.


\section{Training Hyperparameters and Configuration}
\label{app:hyperparams}

All three models are trained with the same optimiser and schedule on 6 NVIDIA H100 GPUs.
Qwen3-8B and OLMo-3-7B use their native tool-calling syntax, while Qwen2.5-7B employs a custom XML tag format (\cref{app:prompts}).
Training uses the verl framework~\citep{sheng2024hybridflow} with FSDP~\citep{zhao2023fsdp} (parameter and optimiser offloading, gradient checkpointing) and SGLang~\citep{zheng2024sglang} for rollout inference.
\Cref{tab:app_hyperparams} lists the full hyperparameter configuration.

\begin{table}[htbp]
  \centering
  \small
  \begin{threeparttable}
    \caption{Training hyperparameters shared across all three models.}
    \label{tab:app_hyperparams}
    \begin{tabular}{lll}
      \toprule
      \textbf{Category} & \textbf{Hyperparameter}                                          & \textbf{Value}                                    \\
      \midrule
      \multirow{7}{*}{Trainer}
                        & Algorithm                                                        & GRPO                                              \\
                        & Learning rate                                                    & $1 \times 10^{-6}$                                \\
                        & LR warmup ratio                                                  & 0.03                                              \\
                        & Gradient clip                                                    & 1.0                                               \\
                        & KL coefficient ($\beta$)                                         & $1 \times 10^{-3}$                                \\
                        & Steps per iteration (per role)                                   & 20                                                \\
                        & Iterations                                                       & 3                                                 \\
      \midrule
      \multirow{7}{*}{Rollout}
                        & Batch size\tnote{$\dagger$}\textsuperscript{,}\tnote{$\ddagger$} & 64 / 256                                          \\
                        & Group size ($G$)                                                 & 8                                                 \\
                        & Max prompt length\tnote{$\dagger$}                               & 1024 / 2048 tokens                                \\
                        & Max response length                                              & 8192 tokens                                       \\
                        & Rollout prompt length                                            & 2048 tokens                                       \\
                        & Max model length                                                 & 16{,}384 tokens                                   \\
                        & Engine                                                           & SGLang                                            \\
      \midrule
      \multirow{8}{*}{Solver Reward}
                        & Rubric weight ($\lambda_{\mathrm{acc}}$)                         & 1.0                                               \\
                        & Format weight ($\lambda_{\mathrm{fmt}}$)                         & 0.5                                               \\
                        & Search weight ($\lambda_{\mathrm{search}}$)                      & 0.1                                               \\
                        & Search reward max turns                                          & 3                                                 \\
                        & Length soft limit ($\ell_{\mathrm{soft}}$)                       & 1024 tokens                                       \\
                        & Length hard limit ($\ell_{\mathrm{hard}}$)                       & 2048 tokens                                       \\
                        & Length penalty floor ($\ell_{\min}$)                             & 0.05                                              \\
      \midrule
      \multirow{4}{*}{Challenger Reward}
                        & Format weight ($\lambda_{\mathrm{fmt}}$)                         & 0.5                                               \\
                        & Difficulty weight ($\lambda_{\mathrm{diff}}$)                    & 1.0                                               \\
                        & Difficulty target ($\tau$)                                       & 0.5                                               \\
                        & Solver rollouts per task ($K$)                                   & 8                                                 \\
      \midrule
      \multirow{4}{*}{Retrieval}
                        & Corpus                                                           & 2018 English Wikipedia~\citep{karpukhin2020dense} \\
                        & Encoding model                                                   & E5-base-v2~\citep{wang2022text}                   \\
                        & Documents per query                                              & 3                                                 \\
                        & Max tool response length                                         & 500 tokens                                        \\
      \midrule
      \multirow{5}{*}{Self-Judge}
                        & Rubric generation temperature                                    & 0.0                                               \\
                        & Rubric generation max tokens                                     & 8192                                              \\
                        & Min rubrics per task                                             & 3                                                 \\
                        & Grading temperature                                              & 0.6                                               \\
                        & Grading max tokens                                               & 16{,}384                                          \\
      \bottomrule
    \end{tabular}
    \begin{tablenotes}
      \footnotesize
      \item[$\dagger$] Challenger / Solver. All other values are shared.
      \item[$\ddagger$] The number of unique prompts per gradient update; each prompt generates $G$ rollouts.
    \end{tablenotes}
  \end{threeparttable}
\end{table}

\paragraph{Qwen3-8B: per-turn datums.}
Qwen3-8B natively produces \texttt{<think>} blocks before each assistant response, including in multi-turn tool-use conversations.
At inference time, prior thinking is not visible to the model on subsequent turns; however, naively constructing training inputs from full trajectories retains all prior thinking blocks, creating a train--inference mismatch.
We split each multi-turn trajectory into per-turn training datums: each datum has thinking trimmed from all previous assistant turns while preserved for the current turn being trained on.
This ensures the training distribution matches the inference distribution at every turn.


\section{Reward Function Details}
\label{app:reward}

This section describes the implementation of each reward component, covering the computational pipeline from raw rollout output to final scalar reward.

\subsection{Challenger Format Reward}
\label{app:reward:challenger}

Each Challenger rollout is decoded into a sequence of assistant turns.  Three components are scored and averaged to produce the format reward $r_{\mathrm{fmt}} \in [0, 1]$:
\begin{enumerate}[leftmargin=*,nosep]
  \item \textbf{Think reward}: the fraction of assistant turns containing a \texttt{<think>}$\ldots$\texttt{</think>} block.  Each assistant turn in the decoded conversation is matched against the regex \texttt{<think>.*?</think>}; the think reward equals $\text{think\_count} / \text{num\_assistant\_turns}$.
  \item \textbf{Tool reward}: the number of assistant turns containing a valid \texttt{<search>}$\ldots$\texttt{</search>} tag (in the \texttt{search\_r1} format), divided by the per-item expected search turn count (sampled from the ``4:3:2'' stratified distribution over 1/2/3 turns).  A search tag is ``valid'' if it is well-formed XML with non-empty query content.  The tool reward is clamped to $[0, 1]$.
  \item \textbf{Structure reward}: 1.0 if the final assistant turn contains a well-formed \texttt{<task>} XML element with the required sub-elements (\texttt{<question>} at minimum), 0.0 otherwise.
\end{enumerate}
The format score is $r_{\mathrm{fmt}} = (\text{think} + \text{tool} + \text{structure}) / 3$.  Rollouts with $r_{\mathrm{fmt}} = 0$ are excluded from difficulty estimation and receive zero total reward.

\subsection{Solver Format and Search Rewards}
\label{app:reward:solver}

\paragraph{Format reward.}
The full multi-turn conversation (prompt + response) is decoded and segmented into assistant turns.  Three components are scored:
\begin{enumerate}[leftmargin=*,nosep]
  \item \textbf{Think reward}: the fraction of assistant turns containing \texttt{<think>}$\ldots$\texttt{</think>} tags.  Every assistant turn is expected to open with a reasoning block; turns without one reduce the think reward proportionally.
  \item \textbf{Tool reward}: the ratio of valid \texttt{<search>} calls in non-final assistant turns to the number of non-final turns, i.e., $\min(\text{valid\_search} / (\text{num\_turns} - 1),\; 1.0)$.  A search call is ``valid'' if it matches the regex \texttt{<search>.*?</search>} with non-empty content.
  \item \textbf{Answer reward}: 1.0 if the final response contains a valid \texttt{<answer>}$\ldots$\texttt{</answer>} block (verified by attempting extraction of the answer text between tags), 0.0 otherwise.
\end{enumerate}
The format score is $F(a) = (\text{think} + \text{tool} + \text{answer}) / 3$.

\paragraph{Search reward.}
The search reward is separate from the tool component of the format score and provides an additive bonus for using retrieval:
\begin{equation*}
  S(a) = \min\!\left(\frac{\text{valid\_search\_calls}}{3},\; 1.0\right),
\end{equation*}
where ``valid\_search\_calls'' counts all well-formed \texttt{<search>} tags across \emph{all} assistant turns (including the final turn, unlike the tool component of the format score which only considers non-final turns).  The denominator of 3 means the reward saturates once the Solver has performed at least 3 retrieval steps, incentivizing multi-hop search without rewarding excessive retrieval beyond the saturation point.

\paragraph{Rubric quality across model scales.}
\Cref{tab:rubric_quality_example} illustrates how rubric specificity varies with model size on a representative summarisation task.

\begin{table}[htbp]
  \centering
  \small
  \begin{tabular}{@{}p{0.12\linewidth}p{0.82\linewidth}@{}}
    \toprule
    \multicolumn{2}{@{}l}{\textit{Task: Summarise events behind David Tuckman's arena project and basketball franchise in Bellevue, WA.}} \\
    \midrule
    \multirow{2}{*}{Qwen3-4B}
     & \rub{1}{Identifies \vague{key events and motivations} of Tuckman's initiative.}                                                    \\
     & \rub{2}{Accurately summarises \vague{financial details and project goals}.}                                                        \\
    \midrule
    \multirow{3}{*}{Qwen3-8B}
     & \rub{1}{Synthesises essential information without omitting the \fact{\$150 million} project and \fact{2001} proposal.}             \\
     & \rub{2}{Connects the arena project to the team's role as \fact{anchor tenant}.}                                                    \\
     & \rub{3}{Correctly attributes the \fact{\$125,000} franchise purchase to Tuckman in \fact{2003}.}                                   \\
    \midrule
    \multirow{3}{*}{Qwen3-32B}
     & \rub{1}{Identifies key events including the \fact{2001} arena proposal and \fact{2003} CBA franchise purchase.}                    \\
     & \rub{2}{Incorporates \fact{\$150 million} arena cost and \fact{\$125,000} acquisition fee.}                                        \\
     & \rub{3}{Explains motivations such as \why{securing long-term tenants for financing} and establishing an \fact{anchor tenant}.}     \\
    \bottomrule
  \end{tabular}
  \caption{Rubrics from Qwen3-4B, 8B, and 32B on a summarisation task.
    \vague{Red}: generic prompt restatements; \fact{blue}: source-grounded facts; \why{green}: analytical requirements.
    4B rubrics omit concrete facts, so generic responses can satisfy them.
    8B and 32B both require specific details from the source document and produce largely overlapping criteria; here, 32B also adds deeper analytical requirements.}
  \label{tab:rubric_quality_example}
\end{table}


\section{Task Generation Pipeline}
\label{app:task_pipeline}

\paragraph{Challenger prompt creation.}
Each iteration generates 2{,}000--3{,}000 challenger training prompts by sampling documents from the corpus and pairing them with task types.
The number of required search turns per prompt is allocated via a stratified ratio (``4:3:2'' for 1/2/3 turns), ensuring diverse retrieval depths.
Prompts are shuffled with a per-iteration seed to ensure non-overlapping data across iterations.

\paragraph{Task creation and filtering.}
After Challenger training (Stage~1), the trained Challenger generates candidate tasks over fresh corpus samples.
Each candidate task undergoes:
\begin{enumerate}[leftmargin=*,nosep]
  \item \textbf{Quality gating}: the Judge evaluates entity identifiability, source relevance.
  \item \textbf{Rubric generation}: the Judge produces 3--5 task-specific rubrics from the task and source document.
  \item \textbf{Difficulty estimation}: $N_{\mathrm{filter}} = 4$ solver rollouts are generated and graded; the mean rubric score $\bar{g}$ is computed.
  \item \textbf{Difficulty filtering}: tasks with $\bar{g} \notin [0.2, 0.8]$ are discarded.
\end{enumerate}
The pipeline targets 5{,}120 filtered tasks per iteration ($256 \text{ batch} \times 20 \text{ steps}$), generating candidate tasks in batches of 200--500 until the target is met.
At most one task is retained per source document (\texttt{max\_tasks\_per\_prompt}=1) to ensure diversity.

Note that the number of solver rollouts differs between stages: Stage~1 Challenger training uses $N_{\mathrm{train}} = 8$ rollouts for difficulty estimation (higher variance reduction during gradient computation), while Stage~2 filtering uses $N_{\mathrm{filter}} = 4$ rollouts (sufficient for the coarse accept/reject decision).








\section{Prompt Templates}
\label{app:prompts}

This section provides the exact prompt templates used during training.
The Challenger and Solver prompts adapt the search-action syntax to each model's native tool-calling convention; all other elements---task emission (\texttt{<task>}), final answers (\texttt{<answer>}), and all Judge prompts---use the same XML format across all three models.
\Cref{tab:prompt_tool_formats} summarises the model-specific search syntax; the full templates below show the Qwen2.5-7B variant.

\begin{table}[htbp]
  \centering
  \small
  \begin{tabular}{@{}ll@{}}
    \toprule
    \textbf{Model} & \textbf{Search action}                                                                \\
    \midrule
    Qwen2.5-7B     & \texttt{<search>}\,\emph{query}\,\texttt{</search>} \quad                             \\[3pt]
    Qwen3-8B       & \texttt{<tool\_call>\{"name":"search", "arguments":\{...\}\}</tool\_call>} \quad      \\[3pt]
    OLMo-3-7B      & \texttt{<function\_calls>search(query="}\emph{...}\texttt{")</function\_calls>} \quad \\
    \bottomrule
  \end{tabular}
  \caption{Model-specific search-action syntax in Challenger and Solver prompts. Only the search call differs; task emission (\texttt{<task>}) and answer submission (\texttt{<answer>}) use XML tags for all models.}
  \label{tab:prompt_tool_formats}
\end{table}

\begin{figure}[htbp]
  \centering
  \begin{promptbox}
    Generate a challenging long-form task using the source document and retrieved information.

    \#\# Source Document\\
    \{source\_document\}

    \#\# Task Type\\
    You must generate a task of type: **\{task\_type\}**\\
    \{task\_description\}

    \#\# Instructions\\
    You will use exactly **\{num\_search\_turns\}** search turns. Each search must be about ONE specific entity that appears by name in the PREVIOUS search results --- not the source document topic. When told your searches are complete, produce the final task.

    The final task MUST require unique facts from ALL search turns --- if any search turn's results could be removed without affecting the answer, the task is invalid.

    \#\# Search Turn Format\\
    <think>Analysis of search results... What new entity or fact to follow next...</think>\\
    <search>your search query</search>

    \#\# Final Task Format (when told searches are complete)\\
    <think>Your final reasoning here...

    Task Prompt Verification:\\
    - [ ] Each search turn contributed a unique essential fact not in the source document\\
    - [ ] The task requires ALL of these facts to answer fully\\
    - [ ] Entities are identifiable (full names + generic qualifier if needed; acronyms expanded)\\
    - [ ] No answer leakage (no specific facts/dates/stats/secondary names)\\
    - [ ] Requires retrieval (cannot be answered from the prompt alone)\\
    - [ ] Specifies output format\\
    - [ ] English only\\
    </think>\\
    <task>The task prompt for the solver. Must withhold searchable information.</task>
  \end{promptbox}
  \caption{Challenger prompt template (Qwen2.5-7B, custom XML format). The Challenger receives a source document and task type, performs multi-turn search, then produces a task. Model-specific search-action variants are listed in \cref{tab:prompt_tool_formats}. Template variables shown in \{braces\}.}
  \label{fig:prompt_challenger}
\end{figure}

\begin{figure}[htbp]
  \centering
  \begin{promptbox}
    Answer the given question. You must conduct reasoning inside <think> and </think> first every time you get new information. After reasoning, if you find you lack some knowledge, you can call a search engine by <search> query </search> and it will return the top searched results between <information> and </information>. You can search up to \{max\_search\_turns\} times, so plan your searches carefully. If you find no further external knowledge needed, you can directly provide the answer inside <answer> and </answer>, without detailed illustrations. For example, <answer> Beijing </answer>.\{budget\_instructions\} Question: \{question\}
  \end{promptbox}
  \caption{Solver prompt template (Qwen2.5-7B, custom XML format). The Solver reasons in \texttt{<think>} blocks, searches via \texttt{<search>} tags, and produces a final answer in \texttt{<answer>} tags. Model-specific search-action variants are listed in \cref{tab:prompt_tool_formats}.}
  \label{fig:prompt_solver}
\end{figure}

\begin{figure}[htbp]
  \centering
  \begin{promptbox}
    You are an expert grader evaluating a user's response to a task prompt against a SINGLE rubric. Be strict: only award credit when the rubric is completely and unambiguously satisfied.

    <prompt>\{prompt\}</prompt>\\
    <response>\{response\}</response>\\
    <rubric>\{rubric\}</rubric>

    Evaluate the response against the rubric above.

    Scoring values (ONLY these two values are allowed):\\
    - 1 = fully and completely meets the rubric with no omissions\\
    - 0 = does not fully meet the rubric (if the response refuses, does not attempt the task, or only partially satisfies the rubric, score 0)

    When in doubt, score 0. Partial credit does not exist.

    STRICT OUTPUT FORMAT --- follow exactly:\\
    1. First, write your reasoning inside a <think> block\\
    2. Then, write exactly one score inside a <score> block: either 0 or 1\\
    3. Output NOTHING else
  \end{promptbox}
  \caption{Grader prompt template. The Judge evaluates a Solver response against a single rubric criterion with strict binary scoring (0 or 1). Each rubric is graded independently.}
  \label{fig:prompt_grader}
\end{figure}

\begin{figure}[htbp]
  \centering
  \begin{promptbox}
    Generate evaluation rubrics for the following task.

    \#\# Task Type\\
    \{task\_type\}

    \#\# Task Prompt\\
    \{task\_prompt\}

    \#\# Context Documents\\
    \{documents\}

    \#\# Instructions

    Generate **3-5 positive rubrics** that measure excellence at the task's core objectives.

    \#\#\# Rubric Writing Rules

    1. **Conciseness**: Each rubric MUST be a SINGLE sentence of at most \~{}25 words. Each rubric contains a single clear requirement --- not compound.

    2. **Command Verb Requirement**: Every rubric MUST begin with a command verb.\\
    \hspace*{1em}Command verbs: Addresses, Demonstrates, Provides, Incorporates, Synthesises, Identifies, Explains, Compares, Evaluates, Integrates, Supports, Applies

    3. **Priority Levels**: Assign exactly one priority to each rubric:\\
    \hspace*{1em}- `critical' - Core requirements; failure means the response is inadequate\\
    \hspace*{1em}- `important' - Significant quality indicators; affects overall quality substantially\\
    \hspace*{1em}- `bonus' - Excellence markers; distinguishes good from exceptional responses

    4. **Discriminative Power**: For each rubric, verify:\\
    \hspace*{1em}- Is it specific to THIS task (not generic)?\\
    \hspace*{1em}- Would it clearly differentiate a good response from a bad one?\\
    \hspace*{1em}- Can it be reliably and objectively judged?

    \#\# Output Format\\
    <rubrics>\\
    \hspace*{1em}<rubric priority="critical|important|bonus">Single-sentence rubric starting with command verb.</rubric>\\
    </rubrics>
  \end{promptbox}
  \caption{Rubric generation prompt template. The Judge generates 3--5 task-specific evaluation rubrics from the task prompt and source document, each beginning with a command verb and assigned a priority level.}
  \label{fig:prompt_rubric}
\end{figure}

\paragraph{Quality gate prompts.}
Two quality gate prompts validate Challenger-generated tasks (\cref{fig:prompt_qg_entity,fig:prompt_qg_relevance}).
All gates output binary scores (0 or 1) using the <think>...<score> format.

\begin{figure}[htbp]
  \centering
  \begin{promptbox}
    You will see ONLY a task prompt. Grade it for entity identifiability.

    Rubric:\\
    Score 1 if a researcher could find the right person, place, work, or concept from what's written --- even if they'd never heard of it before.\\
    Score 0 only if a central entity is genuinely ambiguous --- the name alone refers to many well-known things and the task gives no context to distinguish them.

    Key principle: Context resolves ambiguity. A name that could be ambiguous in isolation often becomes clear from the task itself. ``John Smith'' alone is ambiguous, but ``John Smith's role in the Jamestown colony'' is not. Judge the full task, not names in isolation.

    When to PASS (score 1):\\
    - The task's context (topic, domain, time period, related entities) makes it clear which person/place/work is meant\\
    - Named creative works (books, films, shows, plays) in quotes or with enough context\\
    - Full organization, institution, or place names\\
    - Named historical concepts, legal doctrines, scientific terms, or events\\
    - Well-known acronyms (MIT, NASA, ADA) or acronyms with the full name also provided

    When to FAIL (score 0):\\
    - Any name --- even a full name --- that lacks disambiguating context\\
    - An obscure unexpanded acronym with no full name\\
    - Completely generic references with no proper noun (``the coach'', ``the team'')

    Now grade:\\
    <task\_prompt>\{task\_prompt\}</task\_prompt>
  \end{promptbox}
  \caption{Quality gate: entity identifiability. Ensures the task references entities that can be independently located via search. Few-shot examples included in the full template are omitted here for brevity.}
  \label{fig:prompt_qg_entity}
\end{figure}

\begin{figure}[htbp]
  \centering
  \begin{promptbox}
    You are a strict binary grader. You will see a SOURCE DOCUMENT and a TASK PROMPT.

    Rubric --- Source-Document Relevance:\\
    Score 1 if the task is topically related to the source document --- i.e., the task asks about people, events, places, concepts, or themes that appear in or directly relate to the source document.\\
    Score 0 if the task is clearly unrelated to the source document --- i.e., the task asks about a topic, domain, or entity that has no connection to the source material.

    Rules:\\
    - The task does NOT need to be answerable solely from the source document; it only needs to be topically related.\\
    - Generic or vague tasks that could apply to any document should score 0.\\
    - Tasks that reference specific entities, events, or themes from the source document should score 1.

    Now grade:\\
    <source\_document>\{source\_document\}</source\_document>\\
    <task\_prompt>\{task\_prompt\}</task\_prompt>
  \end{promptbox}
  \caption{Quality gate: source relevance. Ensures the Challenger-generated task is grounded in the provided source document rather than being a generic prompt. Few-shot examples omitted for brevity.}
  \label{fig:prompt_qg_relevance}
\end{figure}

\paragraph{Task type descriptions.}
The Challenger selects from five task types during training (creative writing is defined but excluded):
\begin{itemize}[leftmargin=*,nosep]
  \item \textbf{Long-form QA:} A single question about how/what/why something works, happened, or exists. Explains ONE topic without requiring organised sections.
  \item \textbf{Summarisation:} Condense or transform existing content into shorter/different formats. Requires source material to work from.
  \item \textbf{Planning:} Create actionable plans with concrete steps, timelines, or milestones.
  \item \textbf{Writing:} Compose formal documents with explicit structure (introduction, body, conclusion). Essays, articles, reports that may require citations.
\end{itemize}

\section{Qualitative Analysis}
\label{app:qualitative}

We systematically analyse Challenger and Solver rollouts across iterations and report our findings below.

\definecolor{hlA}{HTML}{1F77B4}  
\definecolor{hlB}{HTML}{D62728}  
\definecolor{hlC}{HTML}{9467BD}  
\definecolor{hlD}{HTML}{E377C2}  
\definecolor{hlE}{HTML}{8C564B}  
\newtcolorbox{challengerbox}[1][]{%
  colback=challengercolor!5!white,
  colframe=challengercolor!80!black,
  fontupper=\footnotesize,
  fonttitle=\footnotesize\bfseries,
  boxrule=0.8pt,
  arc=2pt,
  width=\textwidth,
  title={#1}
}

\newtcolorbox{solverbox}[1][]{%
  colback=solvercolor!5!white,
  colframe=solvercolor!80!black,
  fontupper=\footnotesize,
  fonttitle=\footnotesize\bfseries,
  boxrule=0.8pt,
  arc=2pt,
  width=\textwidth,
  title={#1}
}

\subsection{Challenger Examples}
\label{app:qualitative:challenger}

Iter-1 Challengers often propose tasks answerable from the source document alone.
By iter-3, Challengers use their search turns to retrieve external content and build tasks around it, creating an information asymmetry the Solver must close through its own retrieval.
\cref{fig:qual_c1} shows an example where the iter-3 task depends on retrieved investigation findings, and \cref{fig:qual_c2} shows a case where the iter-3 task introduces a retrieved comparator that implicitly decomposes the problem into multiple retrieval sub-tasks.
Colour-matched spans link retrieved facts to the generated task.

\begin{figure}[ht!]
  \begin{challengerbox}[Difficulty Calibration]
    \textbf{Source document (excerpt):} \emph{``SS El Faro was an American cargo ship that sank on October 1, 2015, during Hurricane Joaquin near the Bahamas. All 33 crew members were lost. The vessel, built in 1975, operated a regular route between Jacksonville, Florida, and San Juan, Puerto Rico\ldots''}

    \smallskip
    \textbf{Retrieved document (excerpt):} \emph{``The \textcolor{hlA}{NTSB investigation} concluded that the captain's decision to sail into the path of Hurricane Joaquin was the \textcolor{hlB}{probable cause} of the sinking. Contributing factors included \textcolor{hlB}{TOTE Maritime's safety management deficiencies}, the vessel's \textcolor{hlC}{open scuttles and unrepaired hull damage}, and the \textcolor{hlC}{absence of an enclosed lifeboat}\ldots''}

    \medskip
    \textbf{Iter-1 Challenger:}\quad
    \emph{``What hurricane was SS El Faro near when it sank, and how many crew members were lost?''}

    \medskip
    \textbf{Iter-3 Challenger:}\quad
    \emph{``Was the loss of SS El Faro better understood as an unavoidable storm casualty or a preventable systems failure? Give a balanced assessment grounded in the incident and its aftermath.''}
  \end{challengerbox}
  \caption{\textbf{The iter-3 Challenger designs tasks that depend on retrieved evidence.}
    The iter-1 task is source-answerable; the iter-3 task requires knowledge from the retrieved document (highlighted) without naming the specific facts the Solver must find.}
  \label{fig:qual_c1}
\end{figure}

\begin{figure}[ht!]
  \begin{challengerbox}[Task Decomposition]
    \textbf{Source document (excerpt):} \emph{``MV Rena was a 3{,}351-TEU container ship. On 5~October 2011, \textcolor{hlA}{navigation errors near Astrolabe Reef} caused it to run aground near Tauranga, New Zealand, \textcolor{hlB}{resulting in an oil spill}\ldots''}

    \smallskip
    \textbf{Retrieved document (excerpt):} \emph{``\textcolor{hlC}{MV Wakashio} was a Japanese-controlled bulk carrier that \textcolor{hlA}{ran aground on a coral reef} off Mauritius in July 2020. It leaked fuel oil, broke apart, and caused what scientists called \textcolor{hlB}{Mauritius's worst environmental disaster}\ldots''}

    \medskip
    \textbf{Iter-1 Challenger:}\quad
    \emph{``Where did MV Rena run aground, and what environmental damage resulted?''}

    \medskip
    \textbf{Iter-3 Challenger:}\quad
    \emph{``Compare the Rena and \textcolor{hlC}{Wakashio} incidents as maritime environmental disasters, explaining how similar groundings led to different questions of responsibility and impact.''}
  \end{challengerbox}
  \caption{\textbf{A retrieved comparator splits the task into implicit sub-problems.}
    The iter-1 task concerns one incident; the iter-3 task adds \textcolor{hlC}{Wakashio}, requiring the Solver to research both cases and synthesise the comparison.}
  \label{fig:qual_c2}
\end{figure}

\subsection{Solver Examples}
\label{app:qualitative:solver}

\begin{figure}[ht!]
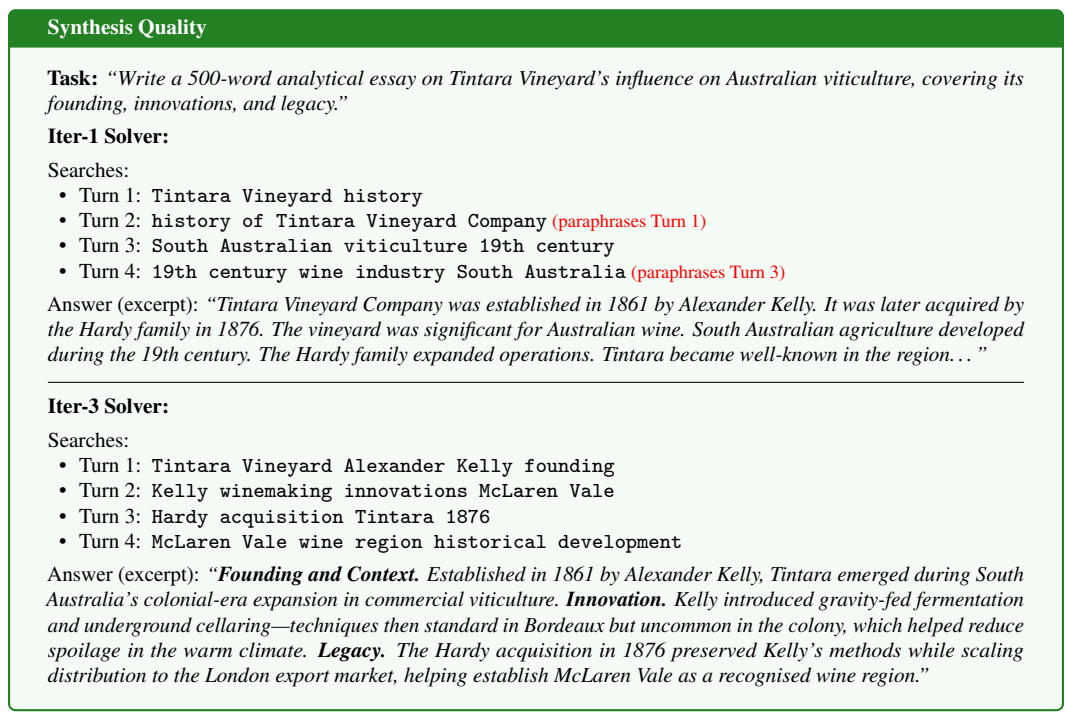

  \begin{solverbox}[Synthesis Quality]
    \textbf{Task:} \emph{``Write a 500-word analytical essay on Tintara Vineyard's influence on Australian viticulture, covering its founding, innovations, and legacy.''}

    \smallskip
    \textbf{Iter-1 Solver:}

    \smallskip
    Searches:
    \begin{itemize}[nosep,leftmargin=1.5em]
      \item Turn 1:\enspace \texttt{Tintara Vineyard history}
      \item Turn 2:\enspace \texttt{history of Tintara Vineyard Company} \textcolor{red}{\scriptsize (paraphrases Turn~1)}
      \item Turn 3:\enspace \texttt{South Australian viticulture 19th century}
      \item Turn 4:\enspace \texttt{19th century wine industry South Australia} \textcolor{red}{\scriptsize (paraphrases Turn~3)}
    \end{itemize}

    \smallskip
    Answer (excerpt): \emph{``Tintara Vineyard Company was established in 1861 by Alexander Kelly. It was later acquired by the Hardy family in 1876. The vineyard was significant for Australian wine. South Australian agriculture developed during the 19th century. The Hardy family expanded operations. Tintara became well-known in the region\ldots''}

    \medskip\hrule\medskip

    \textbf{Iter-3 Solver:}

    \smallskip
    Searches:
    \begin{itemize}[nosep,leftmargin=1.5em]
      \item Turn 1:\enspace \texttt{Tintara Vineyard Alexander Kelly founding}
      \item Turn 2:\enspace \texttt{Kelly winemaking innovations McLaren Vale}
      \item Turn 3:\enspace \texttt{Hardy acquisition Tintara 1876}
      \item Turn 4:\enspace \texttt{McLaren Vale wine region historical development}
    \end{itemize}

    \smallskip
    Answer (excerpt): \emph{``\textbf{Founding and Context.} Established in 1861 by Alexander Kelly, Tintara emerged during South Australia's colonial-era expansion in commercial viticulture. \textbf{Innovation.} Kelly introduced gravity-fed fermentation and underground cellaring---techniques then standard in Bordeaux but uncommon in the colony, which helped reduce spoilage in the warm climate. \textbf{Legacy.} The Hardy acquisition in 1876 preserved Kelly's methods while scaling distribution to the London export market, helping establish McLaren Vale as a recognised wine region.''}
  \end{solverbox}
  \caption{\textbf{Iter-3 organises retrieved facts into coherent, structured analysis.}
    Given the same task, the iter-1 Solver lists facts chronologically; the iter-3 Solver groups them into themes and connects founding context to innovation, and the Hardy acquisition to regional influence.}
  \label{fig:qual_s2}
\end{figure}

Iter-1 Solvers issue generic or repeated queries and often produce flat fact lists.
Iter-3 Solvers use each search turn to pursue a distinct thread, then synthesise the evidence into coherent, structured responses.
\Cref{fig:qual_s1} focuses on retrieval; \cref{fig:qual_s2} focuses on synthesis.

\begin{figure}[ht!]
  \begin{solverbox}[Retrieval Quality]
    \textbf{Task:} \emph{``Why did the Lac-M\'egantic disaster lead to regulatory scrutiny beyond the immediate derailment?''}

    \medskip
    \textbf{Iter-1 Solver:}

    \smallskip
    Searches:
    \begin{itemize}[nosep,leftmargin=1.5em]
      \item Turn 1:\enspace \texttt{Lac-M\'egantic rail disaster}
      \item Turn 2:\enspace \texttt{Lac-M\'egantic train accident cause}
      \item Turn 3:\enspace \texttt{Lac-M\'egantic casualties}
      \item Turn 4:\enspace \texttt{Lac-M\'egantic disaster 2013 Quebec} \textcolor{red}{\scriptsize (paraphrases Turn~1)}
    \end{itemize}

    \smallskip
    Answer (excerpt): \emph{``The Lac-M\'egantic rail disaster was a train accident in Quebec in 2013. An unattended train carrying crude oil derailed and exploded, killing 47 people. The disaster led to investigations and regulatory changes\ldots''}

    \medskip\hrule\medskip

    \textbf{Iter-3 Solver:}

    \smallskip
    Searches:
    \begin{itemize}[nosep,leftmargin=1.5em]
      \item Turn 1:\enspace \texttt{Lac-M\'egantic rail disaster}
      \item Turn 2:\enspace \texttt{Montreal Maine and Atlantic Railway safety record}
      \item Turn 3:\enspace \texttt{Lac-M\'egantic unattended train hand brakes}
      \item Turn 4:\enspace \texttt{crude oil rail transport regulation Canada after 2013}
    \end{itemize}

    \smallskip
    Answer (excerpt): \emph{``The disaster exposed failures beyond the derailment itself. Montreal, Maine and Atlantic Railway had a record of safety violations and operated single-person train crews. The TSB investigation found that insufficient hand brakes on the unattended train were the immediate cause, but Transport Canada subsequently tightened regulations on crude-oil-by-rail, including mandatory two-person crews and upgraded tank car standards\ldots''}
  \end{solverbox}
  \caption{\textbf{Iter-3 search follows a chain from event to mechanism to consequence.}
    The iter-1 Solver stays at the event level, querying the disaster name, cause, and casualty count before paraphrasing its first query.
    The iter-3 Solver uses each turn for a distinct thread: operator practices, failure mechanism, and regulatory aftermath.}
  \label{fig:qual_s1}
\end{figure}

\end{document}